\documentclass[11pt]{article}
\usepackage[a4paper,total={6.5in,9.5in}]{geometry}

\usepackage[english]{babel}
\usepackage[utf8]{inputenc} 
\usepackage[T1]{fontenc}    
\usepackage{microtype}

\usepackage{mathtools}
\usepackage{amsmath}
\usepackage{amsfonts}
\usepackage{amssymb}
\usepackage{bm}

\usepackage[colorlinks=true,linkcolor=blue]{hyperref}
\usepackage[capitalize]{cleveref}
\usepackage{url}           

\usepackage{siunitx}

\usepackage{booktabs}

\usepackage{graphicx}
\usepackage{subcaption}
\usepackage{float}
\graphicspath{{./figures/}}
\DeclareGraphicsExtensions{.pdf}
\def\Y{{\mathcal Y}}
\def\X{{\mathcal X}}

\def\SL{{\mathcal{SL}}}
\def\WL{{\mathcal{WL}}}

\def\P{{\mathbb P}}
\def\R{{\mathbb R}}
\def\N{{\mathbb N}}

\def\1{{\mathbbm 1}}

\def\hY{{\widehat Y}}

\def\OR{{\texttt{OR}}}

\def\TP{{\text{TP}}}
\def\PP{{\text{PP}}}
\def\FP{{\text{FP}}}

\DeclareMathOperator*{\argmax}{arg\,max}
\DeclareMathOperator*{\argmin}{arg\,min}

\usepackage[numbers]{natbib}

\title{Weakly Supervised Learning Significantly Reduces the\\Number of Labels Required for Intracranial\\Hemorrhage Detection on Head CT}

\author{Jacopo~Teneggi\footnote{Department of Computer Science, Johns Hopkins University, Baltimore, MD, 21218}~\footnote{Mathematical Institute for Data Science (MINDS), Johns Hopkins University, Baltimore, MD, 21218} \and Paul H.~Yi\footnote{University of Maryland Medical Intelligent Imaging (UM2ii) Center, Department of Diagnostic Radiology and Nuclear Medicine, University of Maryland School of Medicine, Baltimore, MD, 21201} \and Jeremias~Sulam\footnote{Department of Biomedical Engineering, Johns Hopkins University, Baltimore, MD, 21218}~\footnotemark[2]}

\begin{document}
\date{}
\maketitle

\begin{abstract}
    Modern machine learning pipelines, in particular those based on deep learning (DL) models, require large amounts of labeled data. For classification problems, the most common learning paradigm consists of presenting labeled examples during training, thus providing \emph{strong supervision} on what constitutes positive and negative samples. As a result, the adequate training of these models demands the curation of large datasets with high-quality labels. This constitutes a major obstacle for the development of DL models in radiology---in particular for cross-sectional imaging (e.g., computed tomography [CT] scans)---where labels must come from manual annotations by expert radiologists at the image or slice-level. These differ from examination-level annotations, which are coarser but cheaper, and could be extracted from radiology reports using natural language processing techniques. This work studies the question of \emph{what kind of labels} should be collected for the problem of intracranial hemorrhage detection in brain CT. We investigate whether \emph{image}-level annotations should be preferred to \emph{examination}-level ones. By framing this task as a multiple instance learning (MIL) problem, and employing modern attention-based DL architectures, we analyze the degree to which different levels of supervision improve detection performance. We find that strong supervision (i.e., learning with local image-level annotations) and weak supervision (i.e., learning with only global examination-level labels) achieve comparable performance in examination-level hemorrhage detection (the task of selecting the images in an examination that show signs of hemorrhage) as well as in image-level hemorrhage detection (highlighting those signs within the selected images). Furthermore, we study this behavior as a function of the number of labels available during training. Our results suggest that local labels may not be necessary at all for these tasks, drastically reducing the time and cost involved in collecting and curating datasets.
\end{abstract}

\section{Introduction}
\label{sec:introduction}
Modern Deep Learning (DL) models continue to drive exciting advances across several medical imaging tasks, from image reconstruction and enhancement \cite{ahishakiye2021survey, chen2022ai, alenezi2021geometric, wang2021review, kang2017deep}, to automatic lesion detection and segmentation \cite{cai2020review, giger2018machine, latif2019medical}. DL models for classification and detection are especially desirable for Computer-Aided Diagnosis (CAD) systems in radiology, potentially supporting clinicians in their decision-making by providing a second opinion on subtle cases, or prioritizing the most severe ones \cite{montagnon2020deep, saba2019present, ueda2019technical}. Indeed, recent results indicate that the performance of these machine learning models can be comparable to that of expert physicians in many scenarios \cite{Patel2019-vk, Rajpurkar2018-pj}, and they hold significant promise for the automation of diagnosis, especially in underserved areas where access to radiology expertise might be limited \cite{Kawooya2012-il, choy2018current, langlotz2019will, panesar2020artificial, weissglass2021contextual, attia2021application}.

In this work, we center our attention on the development of DL models for Intracranial Hemorrhage (ICH) detection in head Computed Tomography (CT). In this context, given a new CT scan, the task is to detect the presence of any type of brain hemorrhage. ICH is a potentially life-threatening condition consisting of bleeding inside of the brain which can have several different causes, from trauma to drug abuse \cite{flanders2020construction}. ICH accounts for approximately 10\% to 20\% of all strokes \cite{an2017epidemiology}, and expert radiologists can diagnose ICH from unenhanced head CT scans by analyzing the location, shape, and size of the lesions \cite{flanders2020construction}. The large number of head CT scans produced daily, and the importance of a quick diagnosis for an effective treatment of severe cases, make ICH detection one of the most popular applications of deep learning in radiology thus far \cite{buchlak2022charting}. Many recent works have explored deep learning solutions to different challenges in developing machine learning pipelines for ICH detection, such as the volumetric nature of CT data, the windowing range, and the lack of confidence in \emph{black-box} predictors \cite{yeo2021review, kaka2021artificial, lee2020detection, lee2019explainable}.

At the same time, the development of these high-performing models can be notoriously time-consuming and expensive, largely due to the significant amount of required training data. The most common approach to training DL models for medical imaging classification and detection is \emph{supervised learning}, wherein a collection of images with ground-truth labels are presented to the model. These examples serve the purpose of describing what constitutes a sample from a given class, or how a specific finding looks like in a given image. Naturally, this requires having access to large amounts of labeled data that must be collected by radiologists who manually annotate hundreds or thousands of images---a laborsome and time-consuming process that often results in very high costs \cite{flanders2020construction}.

Some recent research efforts have explored ways of alleviating these limitations. \emph{Semi supervised} learning approaches, for example, extract low-quality labels automatically from clinical notes stored in the Electronic Health Record (EHR) system of a medical institution. The authors in \cite{eyuboglu2021multi} and \cite{tushar2021classification} show how weak labels extracted automatically from clinical reports enable whole-body abnormality detection in PET/CT and body CT, respectively. Although semi supervised learning alleviates the need for large amounts of data with ground-truth annotations, collecting \emph{some} amount of annotated data remains central to training and, importantly, testing these models, and the central aforementioned limitations persist.

In detection problems in particular---where the label of a sample is determined by the presence of a specific finding---it remains unclear \emph{what kind} of labels should be sought after. In the hemorrhage detection problem described above, should ground-truth binary labels be collected for every image in an examination? This can be implemented by labeling an image as \texttt{`1'} if it contains signs of hemorrhage, or \texttt{`0'} otherwise. Or would coarse, examination-level annotations that only indicate the presence of hemorrhage somewhere in the scan (but not in which image) suffice? On the one hand, it is clear that the amount of information in each label decreases as we provide coarser annotation. That is, there might be other findings in a scan (e.g., midline shift effects, external hemotomas, signs of prior surgery, asymmetries) that may be highly correlated with intracranial hemorrhage in the training data. A coarse examination-level binary label may not provide enough information to disambiguate them. At the same time, coarser annotations can lead to huge improvements in data curation time and annotation, since radiologists need only to provide a binary response for each examination.

In this work, we address these fundamental questions using a weakly supervised approach different from semi supervised learning: \emph{Multiple Instance Learning} (MIL) \cite{dietterich1997solving, maron1997framework, weidmann2003two}. In MIL problems, one regards every input as a \emph{bag of instances}, and the label of the bag is determined by the labels of its instances. This framework naturally fits the problem of hemorrhage detection in head CT, since an examination is considered positive (i.e., its coarse, global label is positive) as soon as it contains at least one image with evidence of hemorrhage (i.e., it contains an image with a positive local label). MIL is a particular case of weakly supervised learning, wherein labels are only available for bags (i.e., examinations) instead of instances (i.e., images). By employing a state-of-the-art MIL model \cite{ilse2018attention} that can be trained with either global or local labels, we study whether strong supervision with expensive local labels leads to significantly higher performance in hemorrhage detection in head CT, or whether weak supervision---which is cheaper to obtain---can provide comparable models. 

\subsection*{Summary of contributions}
We show that weakly supervised learning can produce DL models for ICH detection with performance matching that of DL models trained using strong supervision---all while using $\approx 35$-times fewer labels. Furthermore, these weakly supervised models had better generalization on at least one external dataset. Finally, we show that weakly supervised DL models have comparable localization ability of ICH on both the image- and pixel-levels, which is a key feature towards explainability and building trust with clinician end-users. These results inform how data should be collected for this and other similar tasks in radiology, providing a solution to the primary bottleneck in development of high-performing DL models in medical imaging. 

\section{Results}
\label{sec:results}
For a positive head CT scan, we will refer to \emph{examination-level} hemorrhage detection as the task of retrieving the images that contain signs of ICH; and \emph{image-level} hemorrhage detection as the task of highlighting these findings within the retrieved images. We rephrase both examination- and image-level hemorrhage detection as MIL binary classification problems \cite{dietterich1997solving, maron1997framework, weidmann2003two} (see \cref{sec:learning_paradigms} for details on supervised learning and MIL), and evaluate the performance of models trained with local (image-level) annotations and global (examination-level) labels. We refer to the former as a \textit{strong learner} ($\SL$), as it is trained via strong supervision, and \textit{weak learner} ($\WL$) to the latter, since it only uses weak supervision.

\subsection{Datasets} 
\label{sec:datasets}
We train a strong and a weak learner on the RSNA 2019 Brain CT Hemorrhage Challenge dataset \cite{flanders2020construction}, which comprises 21,784 examinations (with a positive rate of 41\%) for a total of 752,803 images (with a positive rate of 14\%).\footnote{For the sake of simplicity, we will refer to the RSNA 2019 Brain CT Hemorrhage Challenge dataset as ``RSNA dataset'', which is available at: \url{https://www.kaggle.com/c/rsna-intracranial-hemorrhage-detection}.} Every image in the RSNA dataset was labeled by expert neuroradiologists with the type(s) of hemorrhage present (i.e., epidural, intraparenchymal, intraventricular, subarachnoid, or subdural). We use 80\% of the data for training and 20\% for validation. Splits were created by random sampling of examinations, rather than images, and the same splits were used for both models in order to guarantee a fair comparison between them. \cref{table:rsna_data} shows the distribution of positive and negative labels---note that while the total \emph{number of images} is the same for each model, the weak learner has access to $\approx 35$-times fewer \emph{total labels}, which is the average number of images in a scan across the dataset.

\begin{table}[t]
    \begin{center}
    \begin{minipage}{\linewidth}
    \caption{\label{table:rsna_data}Number of positive and negative labels in the RSNA dataset for strong and weak learners.}
    \begin{tabular*}{\textwidth}{@{\extracolsep{\fill}}lcccccc@{\extracolsep{\fill}}}
    \toprule
    & \multicolumn{2}{@{}c@{}}{Training} & \multicolumn{2}{@{}c@{}}{Validation}\\
    \cmidrule{2-3}\cmidrule{4-5}
    Learner & Positive labels & Negative labels & Positive labels & Negative labels\\
    \midrule
    full supervision   & 86,295~($\approx$ 14\%) & 515,635~($\approx$ 86\%) & 21,489 ($\approx$ 14\%) & 129,003 ($\approx$ 86\%)\\
    weak supervision   & 7,100~($\approx$ 40\%) & 10,288~($\approx$ 60\%) & 1,776~($\approx$ 40\%) & 2,572~($\approx$ 60\%)\\
    \bottomrule
    \end{tabular*}
    \end{minipage}
    \end{center}
\end{table}

\begin{table}[t]
    \begin{center}
    \begin{minipage}{\textwidth}
    \caption{\label{table:cq500_ctich_data}Number of positive and negative examinations in the CQ500 and CT-ICH datasets, alongside the total number of images contained in the two datasets.}   
    \begin{tabular*}{\textwidth}{@{\extracolsep{\fill}}lccc@{\extracolsep{\fill}}}
    \toprule
    Dataset & Positive examinations & Negative examinations & Total images\\
    \midrule
    CQ500 \cite{chilamkurthy2018deep}   & 212~($\approx$ 49\%)                   & 224~($\approx$ 51\%)                   & 15,156\\
    CT-ICH \cite{hssayeni2020computed}  & 36~($\approx$ 48\%)                    & 39~($\approx$ 52\%)                    & 2,539\\
    \bottomrule
    \end{tabular*}
    \end{minipage}
    \end{center}
\end{table}

In addition to the validation split of the RSNA dataset, we evaluate our resulting models on two external test sets---the CQ500 dataset (436 examinations with a positive rate of 49\%) \cite{chilamkurthy2018deep} and the CT-ICH dataset (75 examinations with a positive rate of 48\%) \cite{hssayeni2020computed,hssayeni2020intracranial,goldberger2000physiobank}.\footnote{The CQ500 dataset is available at: \url{http://headctstudy.qure.ai/dataset}; the CT-ICH dataset is available at: \url{https://physionet.org/content/ct-ich/1.3.1/}.} Table~\ref{table:cq500_ctich_data} shows the distribution of positive and negative examinations in the two external test sets and their total number of images. We note that the CQ500 dataset only provides examination-level labels, while the CT-ICH dataset provides both image-level labels and manual pixel-level segmentations of the bleeds performed by two expert radiologists. Hence, we extend the CQ500 dataset with the ICH bounding box annotations provided for this dataset by three radiologists with varying degree of experience, available in the BHX dataset \cite{reis2020brain,goldberger2000physiobank}.\footnote{The BHX dataset is available at: \url{https://physionet.org/content/bhx-brain-bounding-box/1.1/}.} We include details on the preprocessing of the images for all three datasets in \cref{sec:data_preprocessing}.

\subsection{Attention-based MIL enables training with local or global labels}
We frame the ICH detection task as an MIL binary classification problem. We include a detailed description of the model architectures and their training procedures in \cref{sec:model_architecture_details,sec:training_procedures}, respectively.\footnote{Code to reproduce the experiments in this paper is available at: \url{https://github.com/Sulam-Group/MIL_ICH}.} Here, we briefly describe how state-of-the-art attention-based MIL models \cite{ilse2018attention} enable us to precisely investigate whether classical strong supervision with expensive local labels provides an advantage over weak supervision with cheap global labels.

\begin{figure}[t]
    \centering
    \subcaptionbox{\label{fig:sl_architecture}Pictorial representation of a strong learner $h$ (i.e., $\SL$).}{\includegraphics[width=0.9\linewidth]{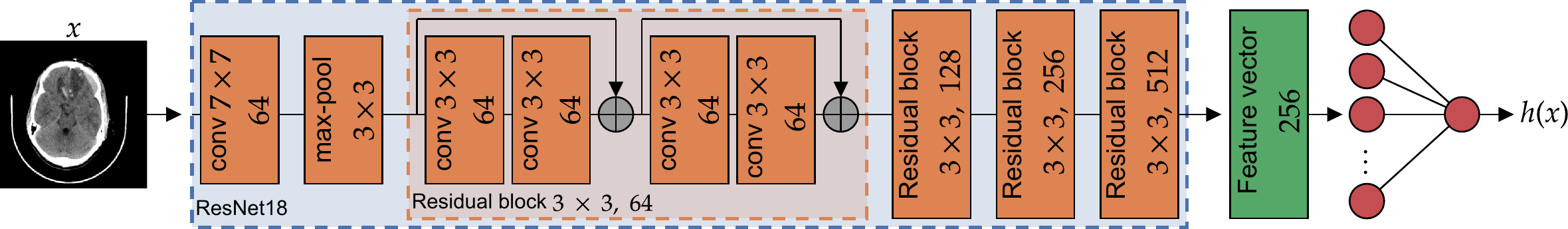}}
    \par\bigskip
    \subcaptionbox{\label{fig:wl_architecture}Pictorial representation of a weak learner $H$ (i.e., $\WL$).}{\includegraphics[width=0.9\linewidth]{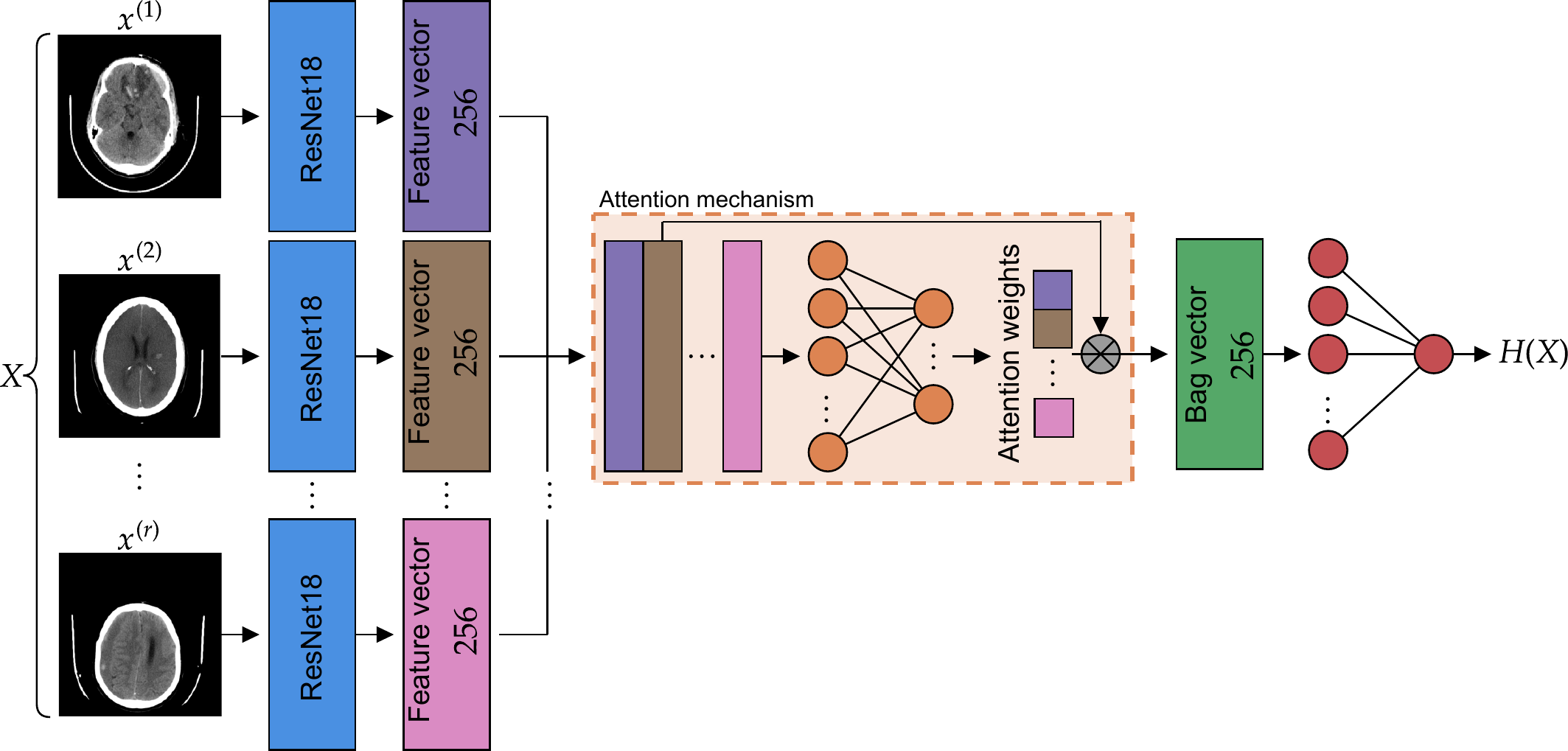}}
    \caption{\label{fig:architectures}Model architectures. \cref{fig:sl_architecture}: The strongly supervised model makes a prediction on every input image, and it requires their labels for training. \cref{fig:wl_architecture}: The weakly supervised model makes a prediction for the entire examination, and thus only requires their labels for training. Note that both learners use the same encoder (i.e., a ResNet18 \cite{he2016deep}) and the same fully-connected layer as the final classifier.}
\end{figure}

We regard an individual image in a CT scan as a vector $x \in \R^d$. Each image is associated with a binary label $y$ that indicates the presence $(y = 1)$ or absence $(y = 0)$ of signs of hemorrhage in the image. A single CT scan of a patient can be naturally seen as the stacking of $r$ images along the scanner's axis, i.e. $X = [x^{(1)}, x^{(2)}, \dots, x^{(r)}] \in \R^{dr}$. Analogously to images, examinations are also associated with a binary label $Y$ indicating whether any image in the examination presents signs of hemorrhage $(Y = 1)$, or every image in the examination is healthy $(Y = 0)$. Note that the label of an examination can be determined from the labels of the images in the examination (if they are available), since the presence of hemorrhage in any image implies the presence of hemorrhage in the examination. This observation can be formalized by stating that the examination's label is the logical $\OR$ function of the labels of the images in the examination, i.e. $Y = \OR(y^{(1)}, y^{(2)}, \dots, y^{(r)})$.

In the traditional strongly supervised setting, a predictor $h$ is trained on a collection of $n_s$ training samples $\{(x_i,y_i)\}_{i=1}^{n_s}$, with the goal of obtaining $h(x)$ as an accurate predictor of the label of a new sample $x$, i.e. as an approximation of the conditional expectation of $y$ given $x$. In this work, $h$ is given by the composition of a feature extractor $f$, implemented by a Convolutional Neural Network (CNN) that encodes a $d$-dimensional input (here $d = 512 \times 512$ pixels) into a feature vector of size 256, with a binary classifier $g$ that receives the feature vector and returns a value in the unit interval $[0, 1]$.  To summarize, $h:~\R^{d} \to [0, 1]$ such that $h(x) = g(f(x))$. We remark that training $h$ requires the collection, annotation, and curation of $n_s$ pairs of input images with their respective labels, which is time-consuming and costly.

To grant these functions the ability to learn with global labels only, we propose an attention-based MIL DL architecture \cite{ilse2018attention} that can predict the presence of hemorrhage on entire examinations of arbitrary length $r$ directly, which we denote $H:~\R^{dr} \to [0, 1]$. This predictor accepts an entire stack of images as input and it predicts the presence or absence of ICH in it. Unlike the previous case, training such a predictor $H$ only requires collecting $n_w$ training samples of pairs $\{(X_i,Y_i)\}_{i=1}^{n_w}$, where $Y_i$ are the global labels of the examinations---and thus, the local labels of each image are not needed. Since there are a large number of images per examination ($r$ is about 30 for a typical scan), the number of examination labels is much lower than that of image labels, i.e. $n_w \ll n_s$.

The predictor $H$ is a multiple instance learning (MIL) model, as it receives as input a collection (i.e., a bag) of $r$ images (i.e., instances). MIL has a long tradition in machine learning \cite{dietterich1997solving,sabato2009homogeneous,sabato2010reducing,sabato2012multi} and in biomedical imaging in particular \cite{amores2013multiple,quellec2017multiple,cheplygina2019not,Wang_MIL_2022}. However, its applications to the task of ICH detection remain limited \cite{saab2019doubly,wu2021combining,lopez2022deep}. Similarly to previous works \cite{saab2019doubly,wu2021combining,lopez2022deep} we use the attention-based MIL framework recently developed in \cite{ilse2018attention}, which parametrizes such an MIL predictor $H$ by composing $r$ instance-wise encoders with an attention mechanism \cite{bahdanau2014neural,vaswani2017attention}, $a$, and a final classifier $g$. Succinctly, we can write $H(X) = g(a( [f(x^{(1)}),f(x^{(2)}),\dots,f(x^{(r)})]))$. This MIL predictor, as well as the strongly supervised one, are depicted in \cref{fig:architectures}.

Importantly, to make comparisons between the local (strong) and global (weak) predictors, the feature extractor $f$ that encodes each image in an examination, as well as the binary classifier $g$, in $H$ are the same as the ones described in the context of traditional supervised learning and used in the image-level predictor $h$. Moreover, if an examination $X$ contains a single image (i.e., $r = 1$) the attention mechanism $a$ reduces to the identity map. It follows that for these cases $H(X) = g(f(X)) = h(x)$, and thus $H$ is equivalent to the fully supervised predictor $h$. For this reason, the MIL model $H$ generalizes the image-wise predictor $h$ while maintaining the core feature extractor and classifier ($f$ and $g$, respectively). In this work, we compare the resulting image-wise classifier $h$, trained using the local annotations from every image, and examination-wise classifier $H$, trained using only global labels for every examination. 

\subsection{MIL provides comparable performance on examination-level binary classification}
We compare the strong and weak learners on the examination-level binary classification problem, i.e. the task of predicting whether a new examination $X$ (with $r$ images) contains any signs of hemorrhage. For the MIL learner $H$, the examination-level prediction is simply $\hY_w = H(X)$. On the other hand, the strongly supervised predictor $h$ can predict on single images only. Since the ground-truth examination-level label $Y$ can be expressed as the logical $\OR$ of the labels of the images in the examination, it is natural to define the examination-level prediction of $h$ as $\hY_s = \max(h(x^{(1)}), h(x^{(2)}), \dots, h(x^{(r)}))$, which extends the logical $\OR$ to real-valued functions on the unit interval $[0, 1]$.

\begin{figure}[t]
    \centering
    \subcaptionbox{\label{fig:global_RSNA_roc}RSNA dataset\\\centering$(p = 0.636)$.}{\includegraphics[width=0.32\linewidth]{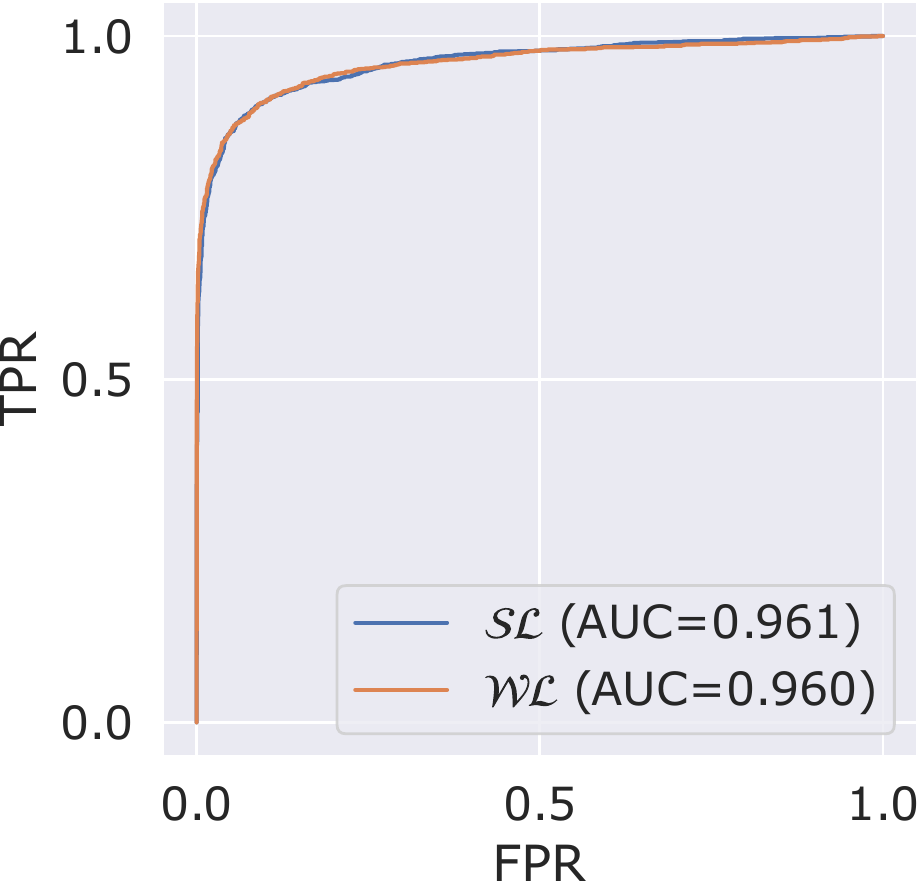}}
    \subcaptionbox{\label{fig:global_CQ500_roc}CQ500 dataset\\\centering$(p = 0.147)$.}{\includegraphics[width=0.32\linewidth]{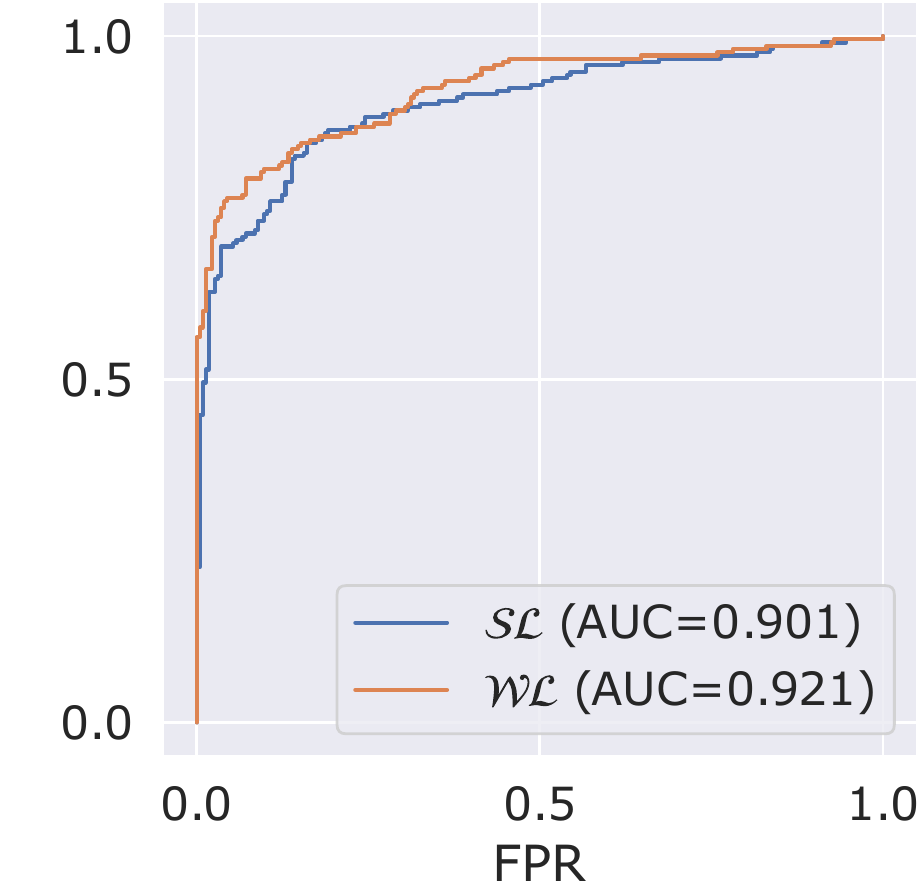}}
    \subcaptionbox{\label{fig:global_CT-ICH_roc}CT-ICH dataset\\\centering$(p = 0.032)$.}{\includegraphics[width=0.32\linewidth]{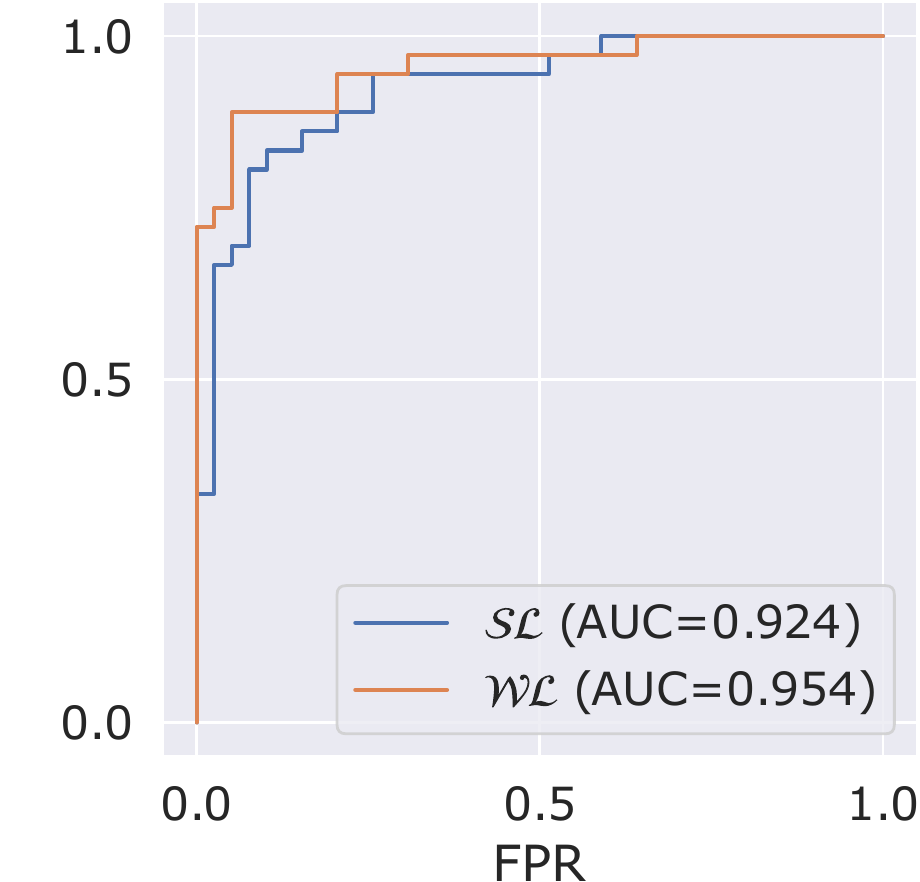}}
    \caption{\label{fig:global_all_roc}Comparison of a strong learner ($\SL$) and an MIL learner ($\WL$) on the examination-level binary classification problem. AUC's are compared via a one-sided DeLong's test.}
\end{figure}

\cref{fig:global_all_roc} shows the ROC curves with their AUC's for the strong learner $h$ (i.e., $\SL$: \emph{strong learner}) and the MIL learner $H$ (i.e., $\WL$: \emph{weak learner}) on the validation split of the RSNA dataset as well as the CQ500 and CT-ICH datasets. AUC's are compared via a one-sided DeLong's test \cite{delong1988comparing}. \cref{fig:global_RSNA_roc,fig:global_CQ500_roc} demonstrate that there is virtually no difference in performance between the strong and the weak learner on the validation split of the RSNA dataset and the CQ500 dataset, respectively. The strong learner achieves an AUC of $0.961$, whereas the weak learner obtains an AUC of $0.960$ ($p = 0.636$) on the validation split of the RSNA dataset, and of $0.901$ and $0.921$ ($p = 0.147$), respectively, on the CQ500 dataset. In fact, \cref{fig:global_CT-ICH_roc} suggests that the weak learner has significantly better generalization power on the CT-ICH dataset (AUC's of $0.924$ and $0.954$, respectively, $p = 0.032$). 

\subsection{MIL provides comparable performance on examination-level hemorrhage detection}
Recall that we refer to \emph{examination-level} hemorrhage detection as the task of retrieving the positive images within a positive examination. That is, identifying those images in a scan that show signs of hemorrhage (if any are present). For the strong learner $h$, this is no different than predicting the presence of hemorrhage individually on each of the images in the examination, and selecting the predicted positive images. For the MIL learner $H$, on the other hand, there is no unique way to perform this image-wise selection. A very popular approach relies on employing the attention mechanism $a$ as an instance selector, since this function explicitly assigns weights (between 0 and 1) to each instance in the bag, thus reflecting some notion of importance towards the overall label of the bag \cite{saab2019doubly,lee2019explainable,schlemper2019attention,roscher2020explainable}. Alternatively, one can resort to other notions of importance, such as those based on game-theoretic principles. Shapley coefficients provide a natural way to do this by assigning scores to each image that reflect their contributions towards the overall examination prediction \cite{Shapley1953,lundberg2017unified,teneggi2022fast}. We explore both of these approaches here.

Intuitively, we expect an accurate MIL predictor $H$ to assign large attention weights to the positive images within a positive examination. Then, we select those images whose attention weights are no smaller than a certain threshold, $t$. In this work we use $t = 1/r$, which corresponds to uniform attention across all $r$ images in an examination, but this choice for $t$ is not crucial and other options exist.\footnote{Indeed, since we employ \textit{sparse-max} \cite{martins2016softmax,peters2019sparse,correia19adaptively} for the attention mechanism, this behavior is relatively independent of the chosen threshold.} Although attention weights are extensively used in recent literature \cite{saab2019doubly,lee2019explainable,schlemper2019attention,roscher2020explainable} to select important instances, their theoretical underpinnings remain scarce \cite{jain2019attention,ethayarajh2021attention}. On the contrary, the Shapley value \cite{Shapley1953} has gained substantial popularity in the machine learning literature \cite{covert2021explaining} because of its precise theoretical properties. Here, we introduce the first Shapley-based explanation method specifically designed for deep set predictors \cite{zaheer2017deep} (such as the MIL predictor $H$) by extending \emph{h-Shap} \cite{teneggi2022fast}, a hierarchical extension of the Shapley value (see \cref{sec:exam_level_hshap} for details).\footnote{h-Shap is available at \url{https://github.com/Sulam-Group/h-shap}.} Similarly to the attention-based selection method, we select the images which have a Shapley value no smaller than $t = 1/r$.

\begin{figure}[t]
    \centering
    \subcaptionbox{\label{fig:exam_level_RSNA_TPR}RSNA dataset}{\includegraphics[width=0.47\linewidth]{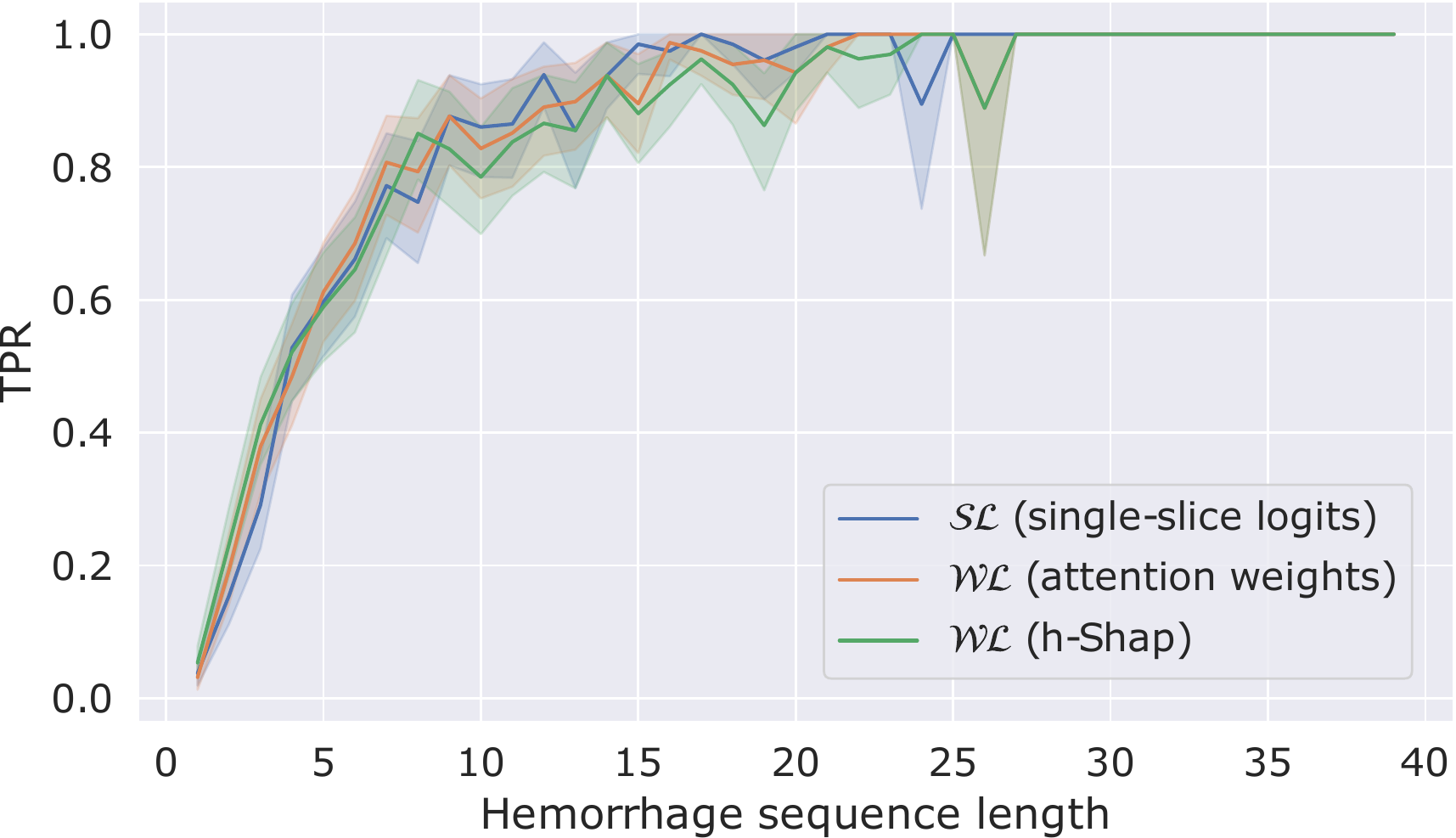}}
    \hfill
    \subcaptionbox{\label{fig:exam_level_CT-ICH_TPR}CT-ICH dataset}{\includegraphics[width=0.47\linewidth]{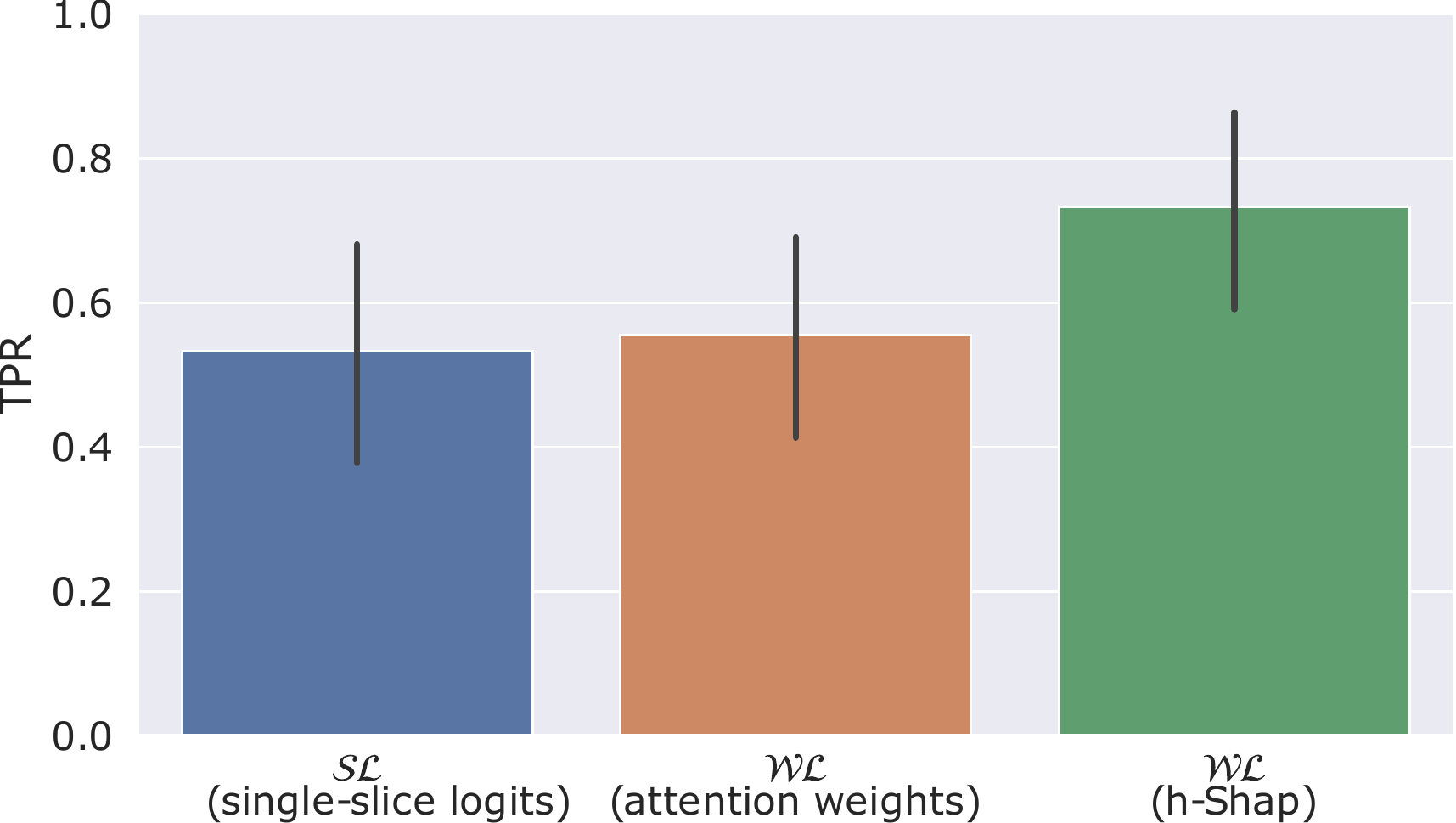}}
    \captionsetup{subrefformat=parens}
    \caption{\label{fig:exam_level_all_TPR}Comparison of a strong learner ($\SL$) and a weak learner ($\WL$) on examination-level hemorrhage detection. \subref{fig:exam_level_RSNA_TPR} Average recall (TPR) as a function of hemorrhage sequence length on the validation split of the RSNA dataset. \subref{fig:exam_level_CT-ICH_TPR} Average TPR on the CT-ICH dataset.}
\end{figure}

We compare the strong and the weak learners on examination-level hemorrhage detection by means of their examination-level $f_1$ score over the true hemorrhage sequences (i.e., series of consecutive positive images) in true positive examinations (see \cref{sec:exam_level_detection_comparison} for details). \cref{fig:exam_level_RSNA_TPR} shows the average recall on the RSNA validation split as a function of the number of consecutive positive images in a hemorrhage sequence. We can appreciate how there is no significant gap between the performance of the strong learner $\SL$ compared to the weak learner $\WL$, with either detection strategy. \cref{fig:exam_level_RSNA_TPR} shows that, as expected, it is in general harder for all learners to detect short hemorrhage sequences that may comprise only a few consecutive positive images. \cref{fig:exam_level_CT-ICH_TPR} shows the average recall on the CT-ICH dataset. In this case, we do not stratify the results as a function of hemorrhage sequence length given the relatively small amount of examinations in the dataset. We see here as well that there in no significant generalization power difference across learners or detection strategies, with a slight advantage for the weak learner with the Shapley-based selection method.

\begin{figure}[t]
    \centering
    \begin{subfigure}{\linewidth}
        \includegraphics[width=0.18\linewidth]{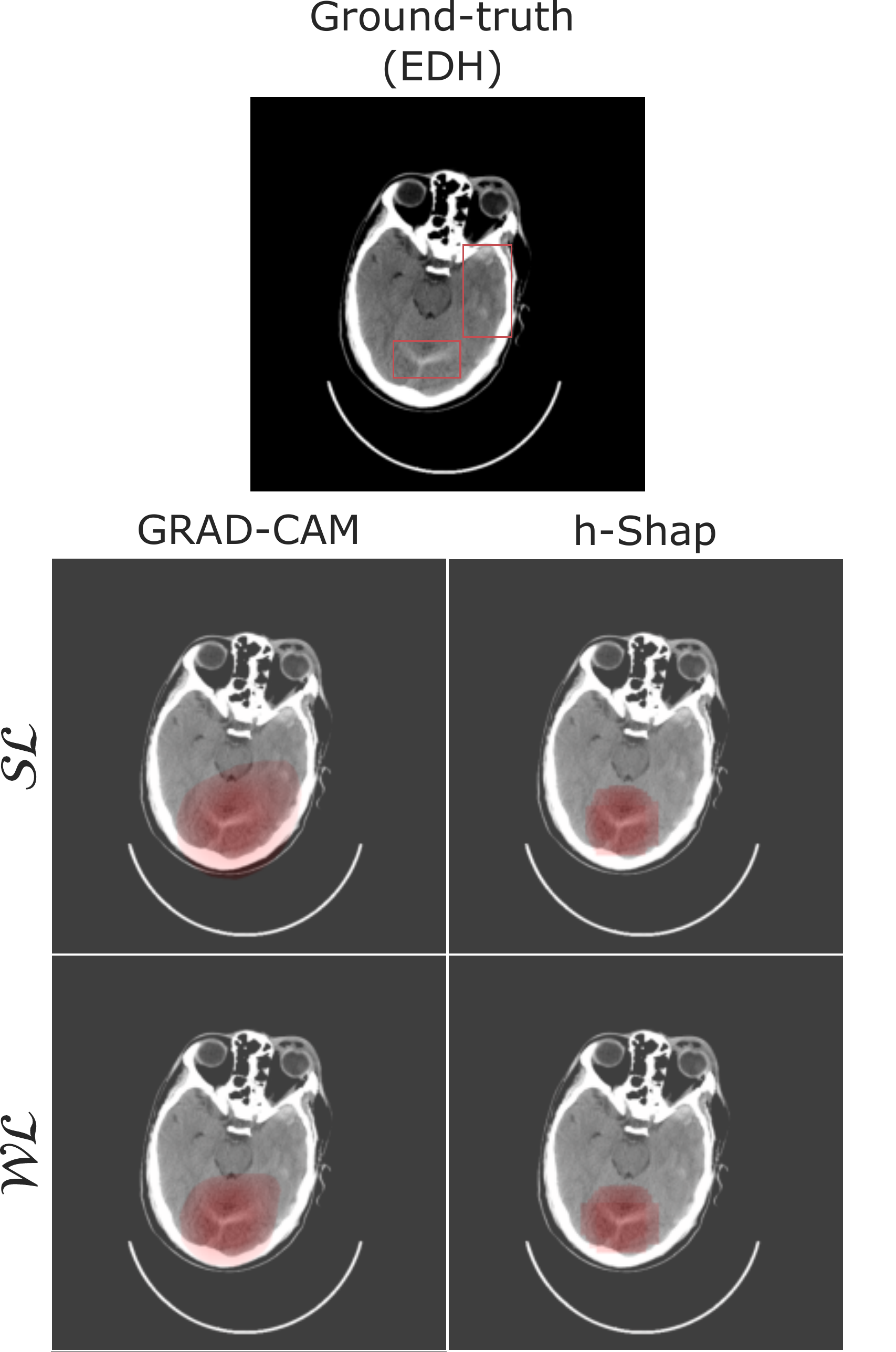}
        \includegraphics[width=0.18\linewidth]{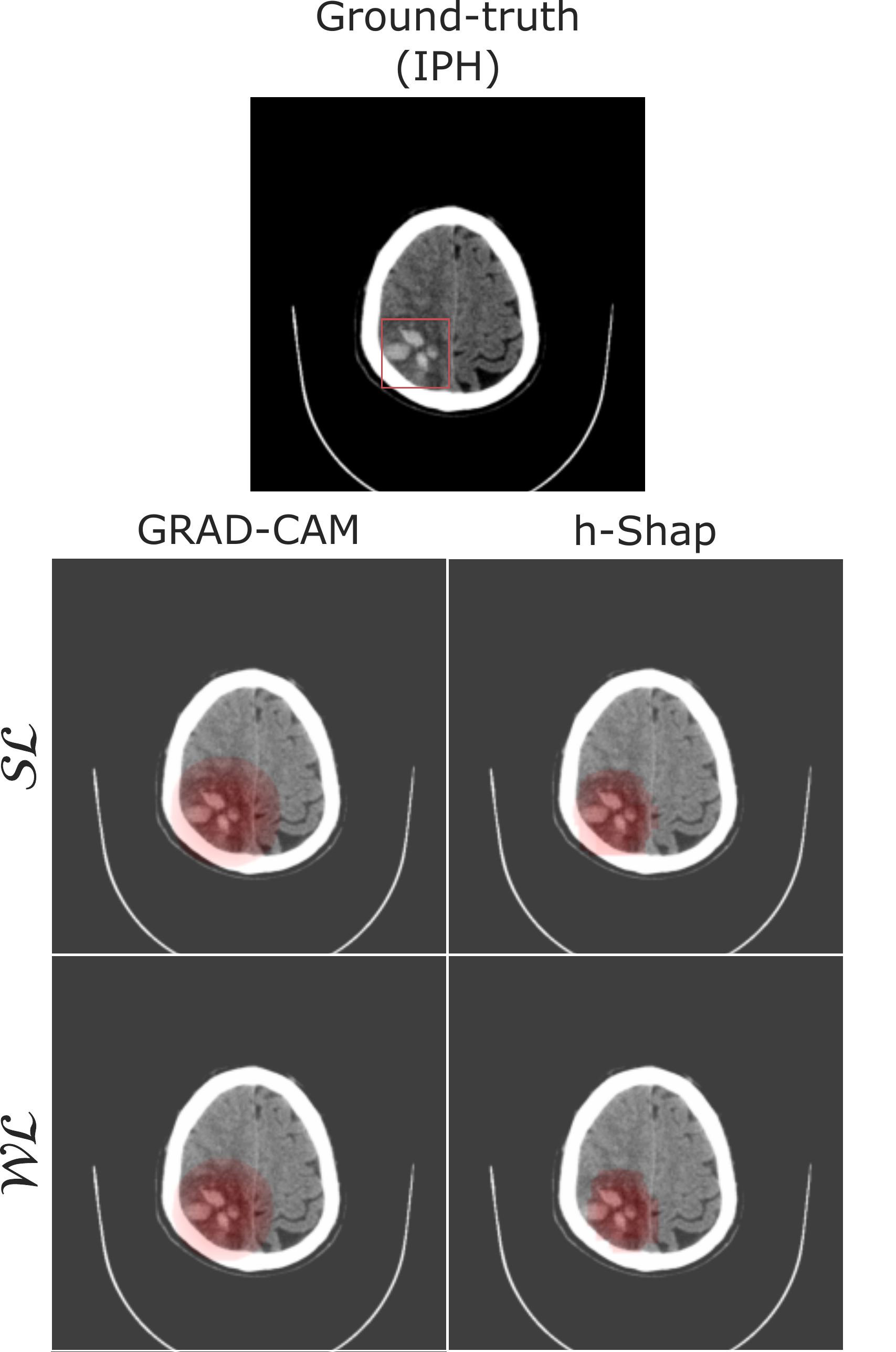}
        \includegraphics[width=0.18\linewidth]{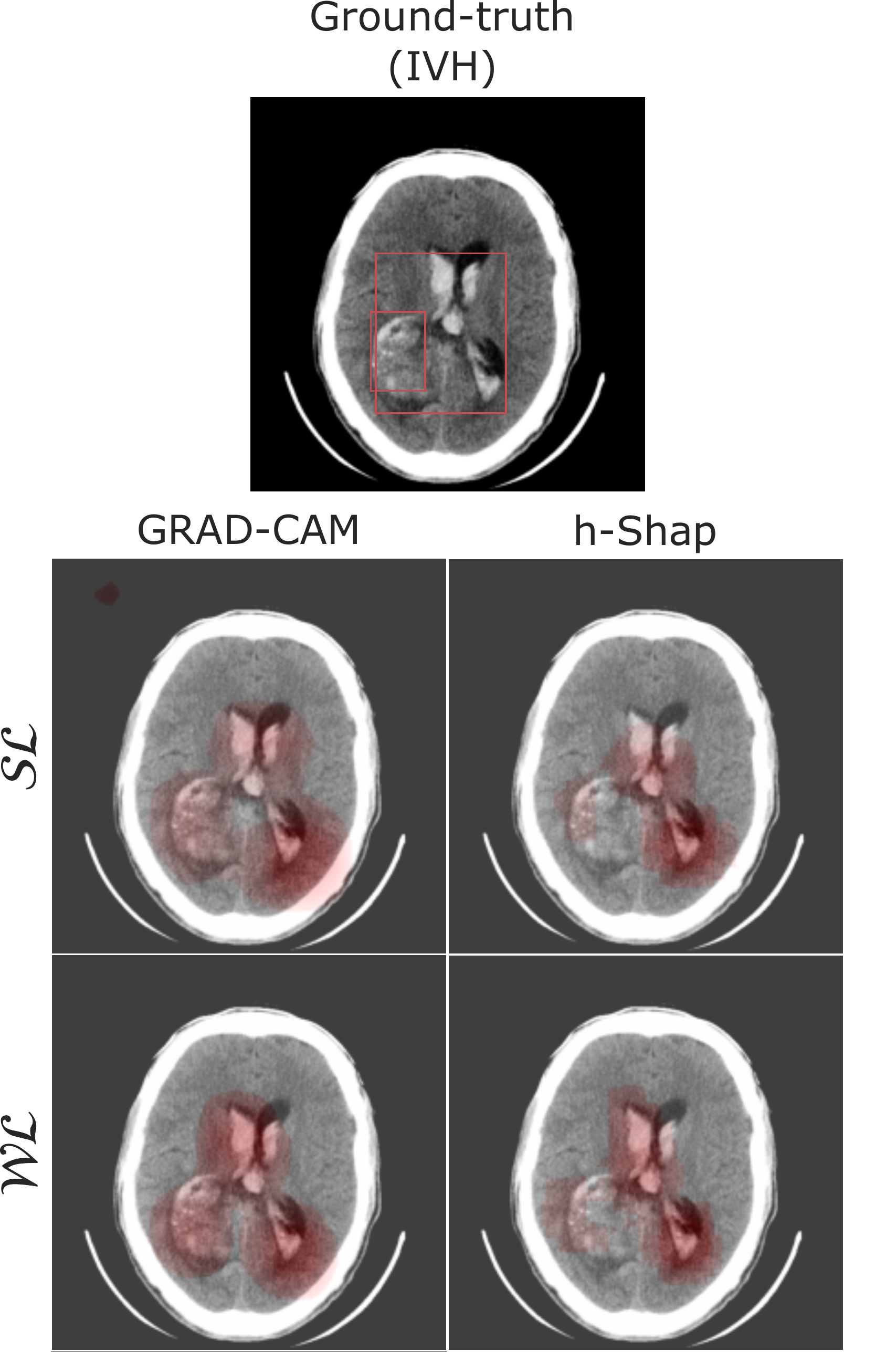}
        \includegraphics[width=0.18\linewidth]{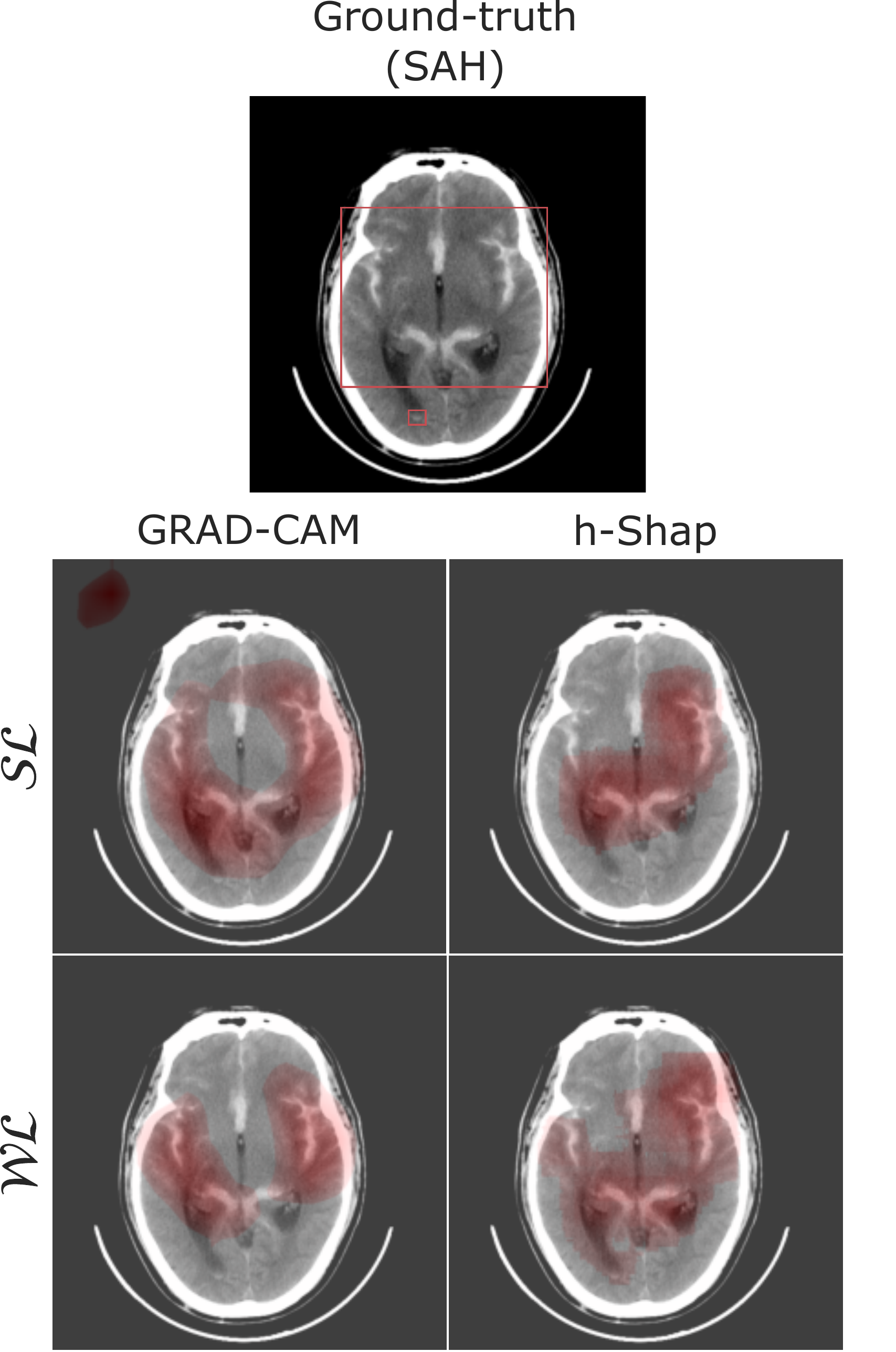}
        \includegraphics[width=0.18\linewidth]{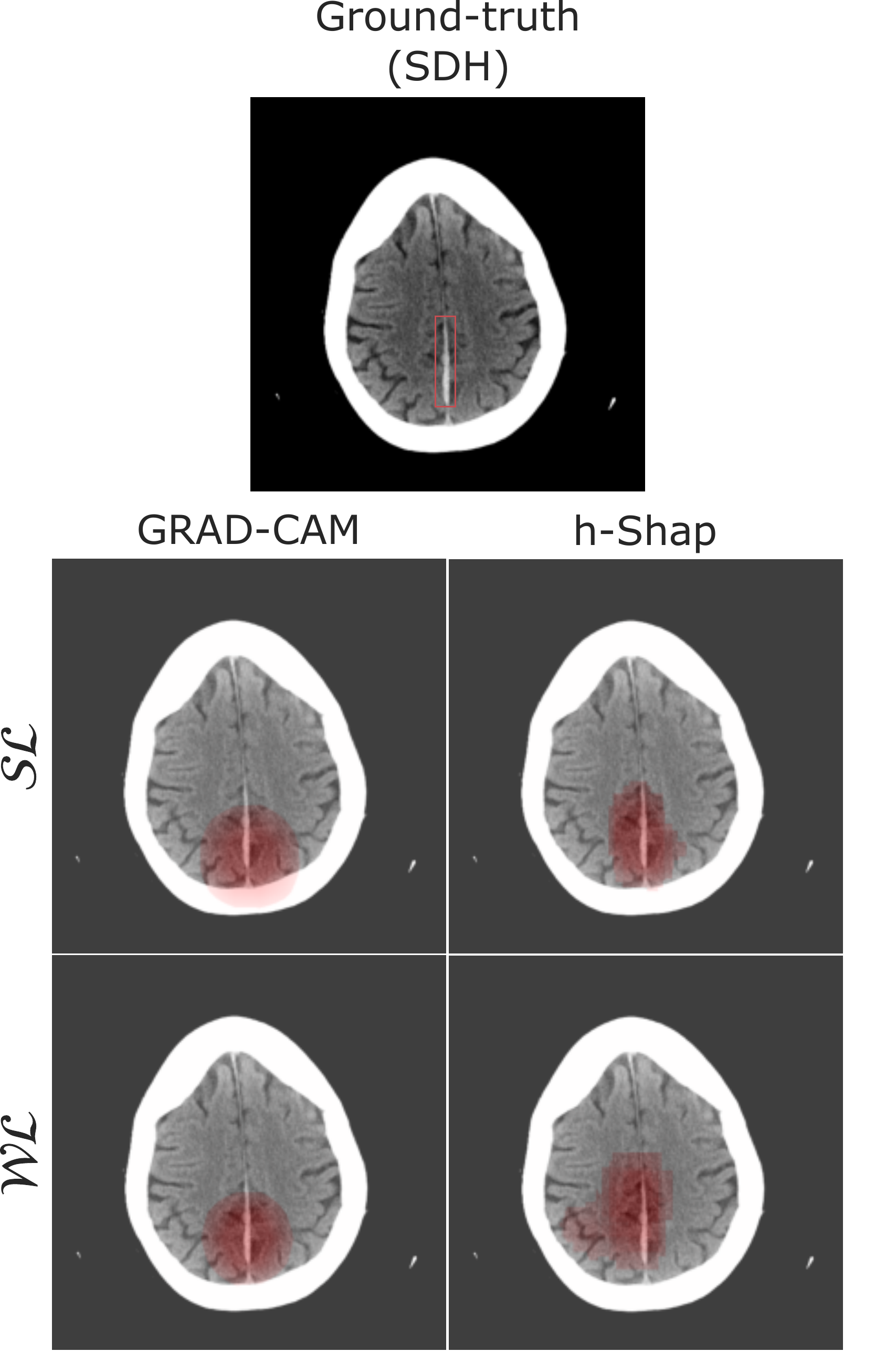}
        \caption{CQ500 dataset.}
    \end{subfigure}
    \begin{subfigure}{\linewidth}
        \includegraphics[width=0.18\linewidth]{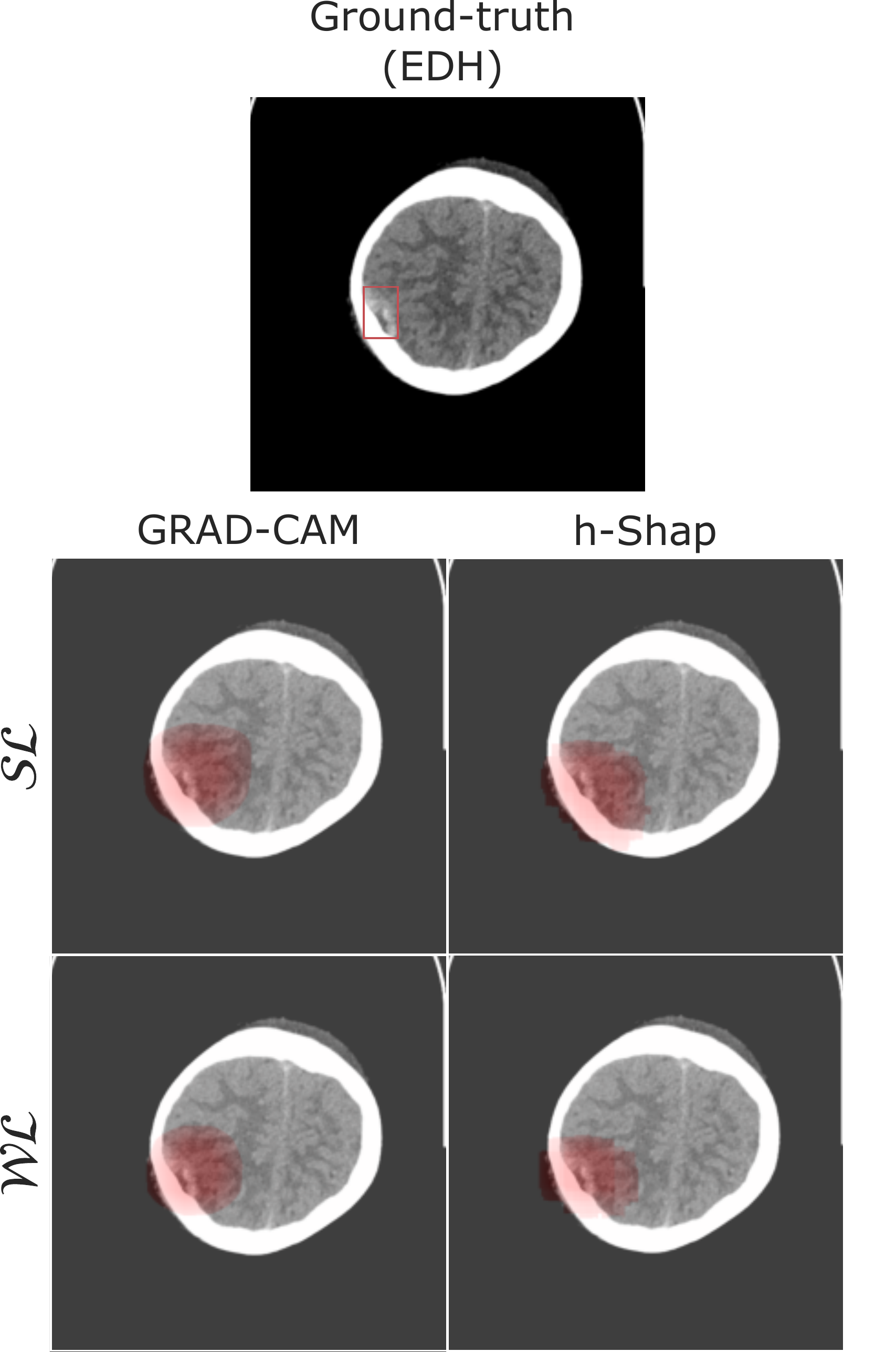}
        \includegraphics[width=0.18\linewidth]{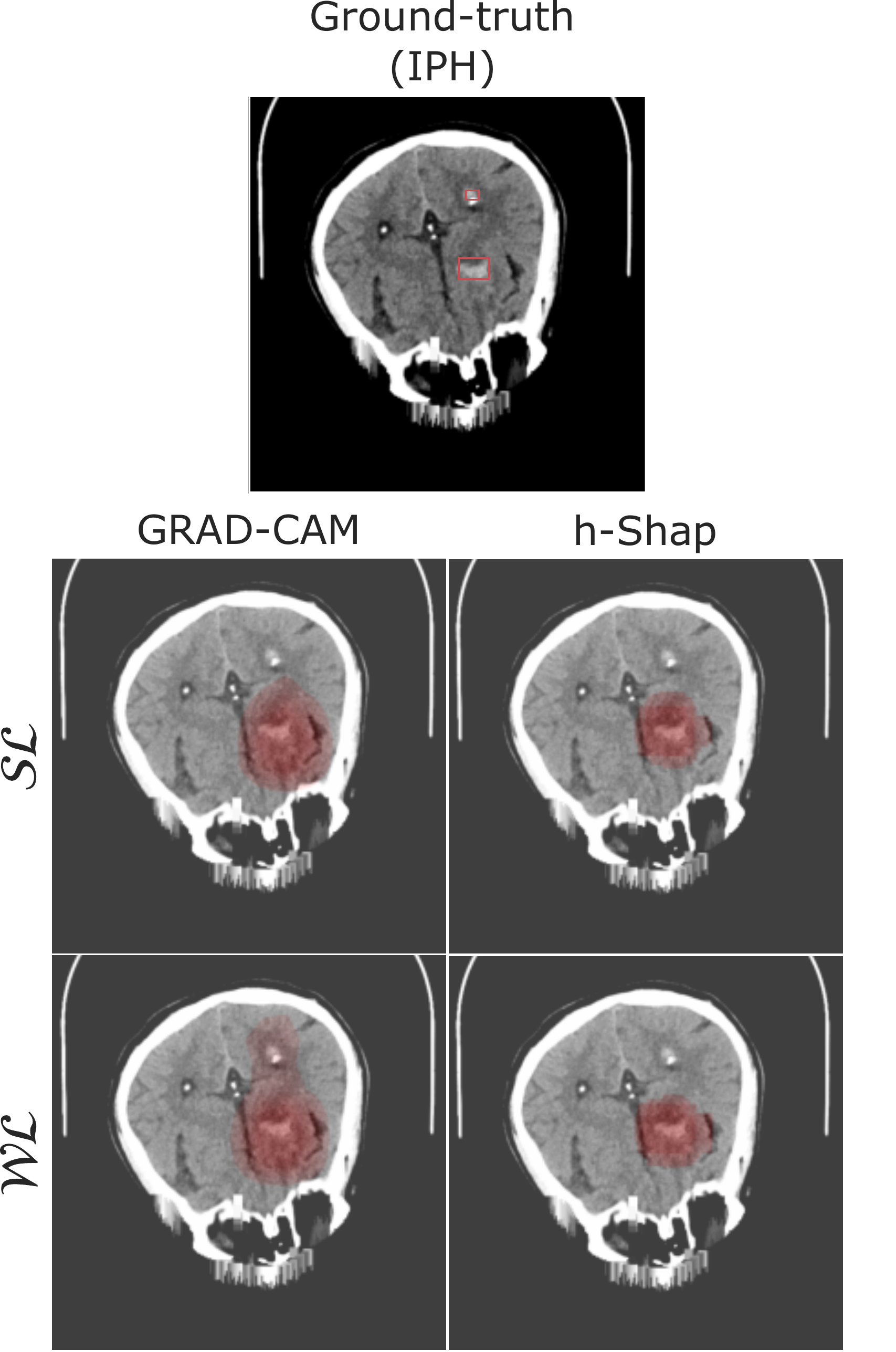}
        \includegraphics[width=0.18\linewidth]{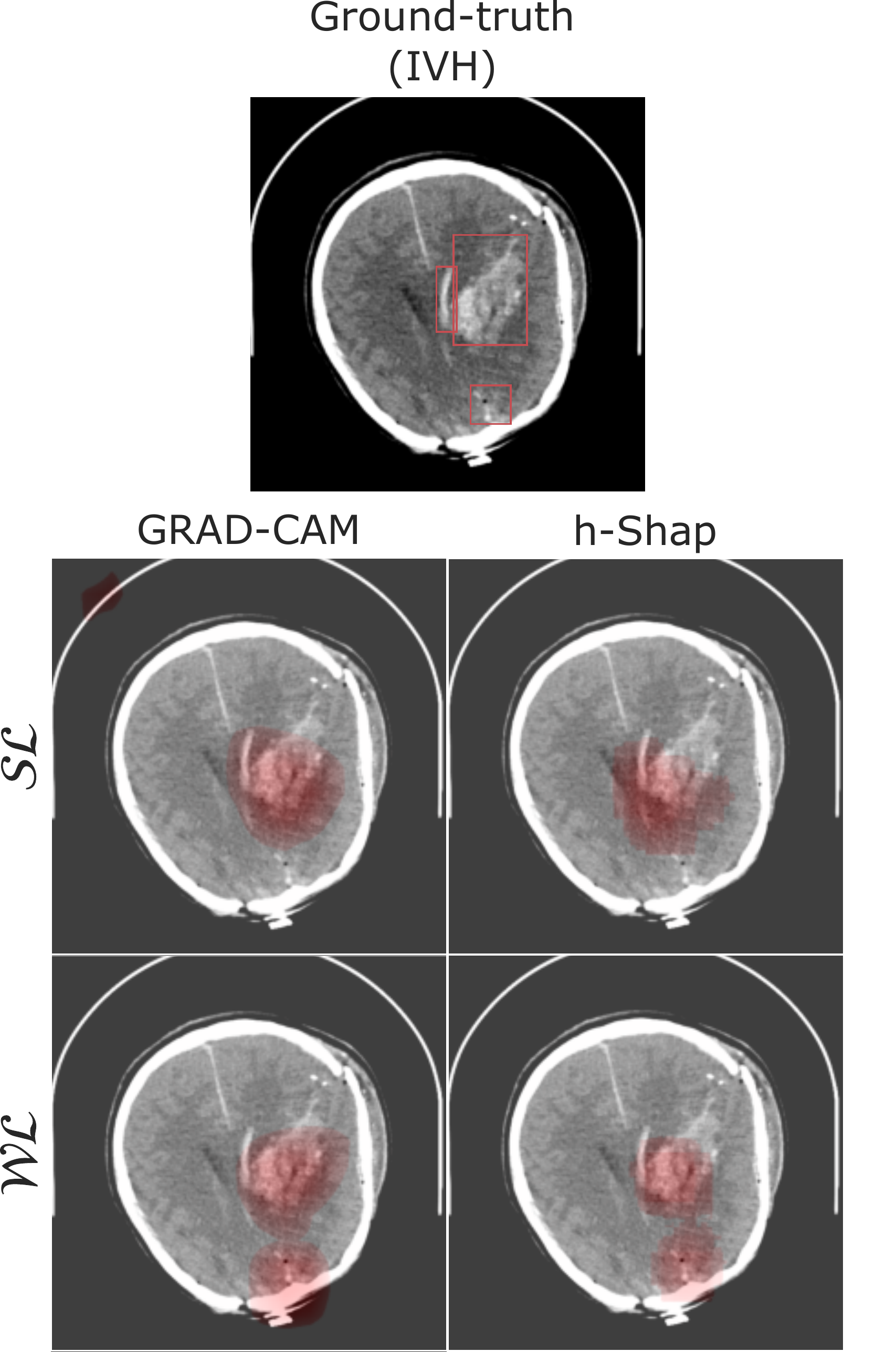}
        \includegraphics[width=0.18\linewidth]{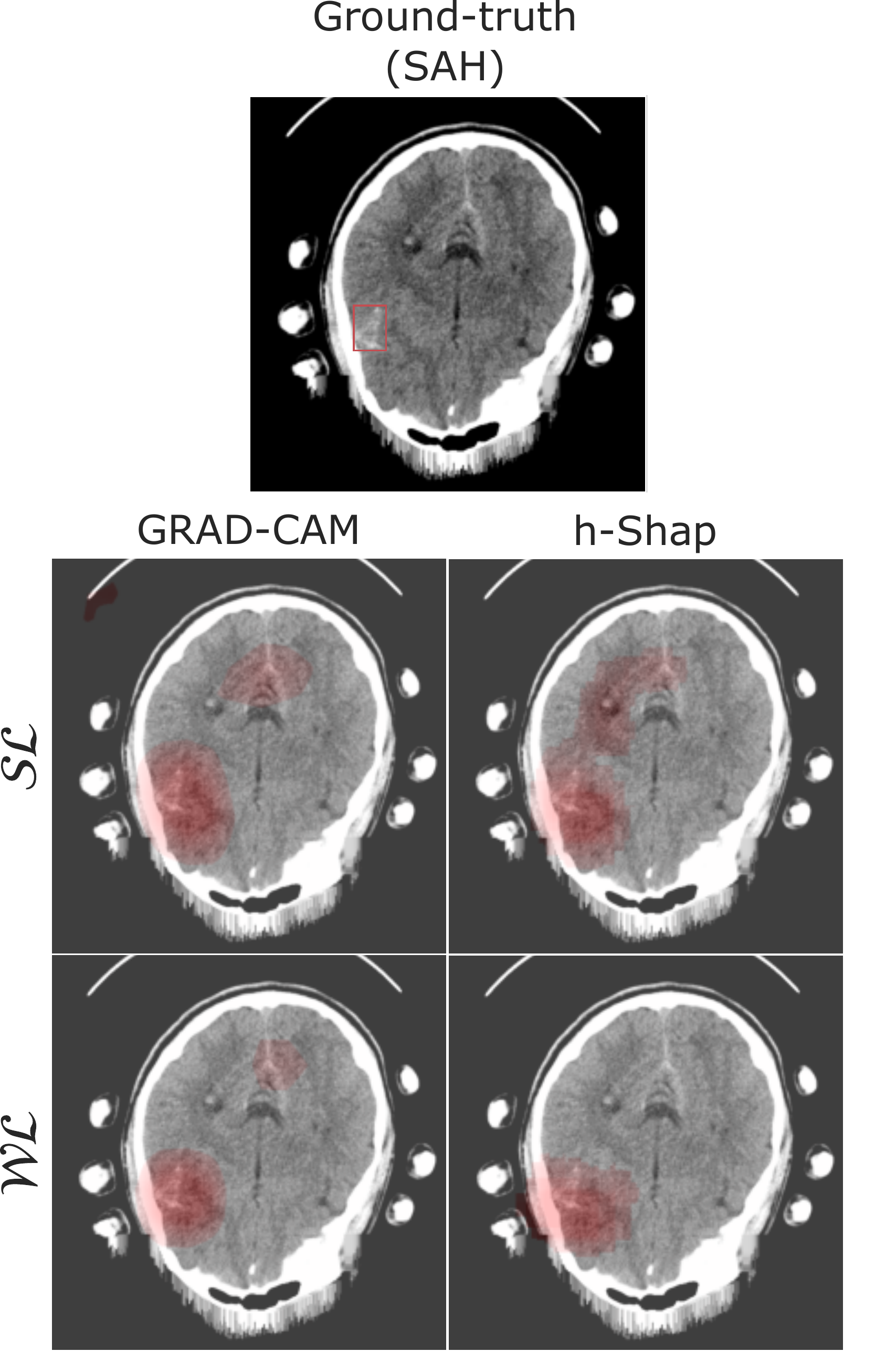}
        \includegraphics[width=0.18\linewidth]{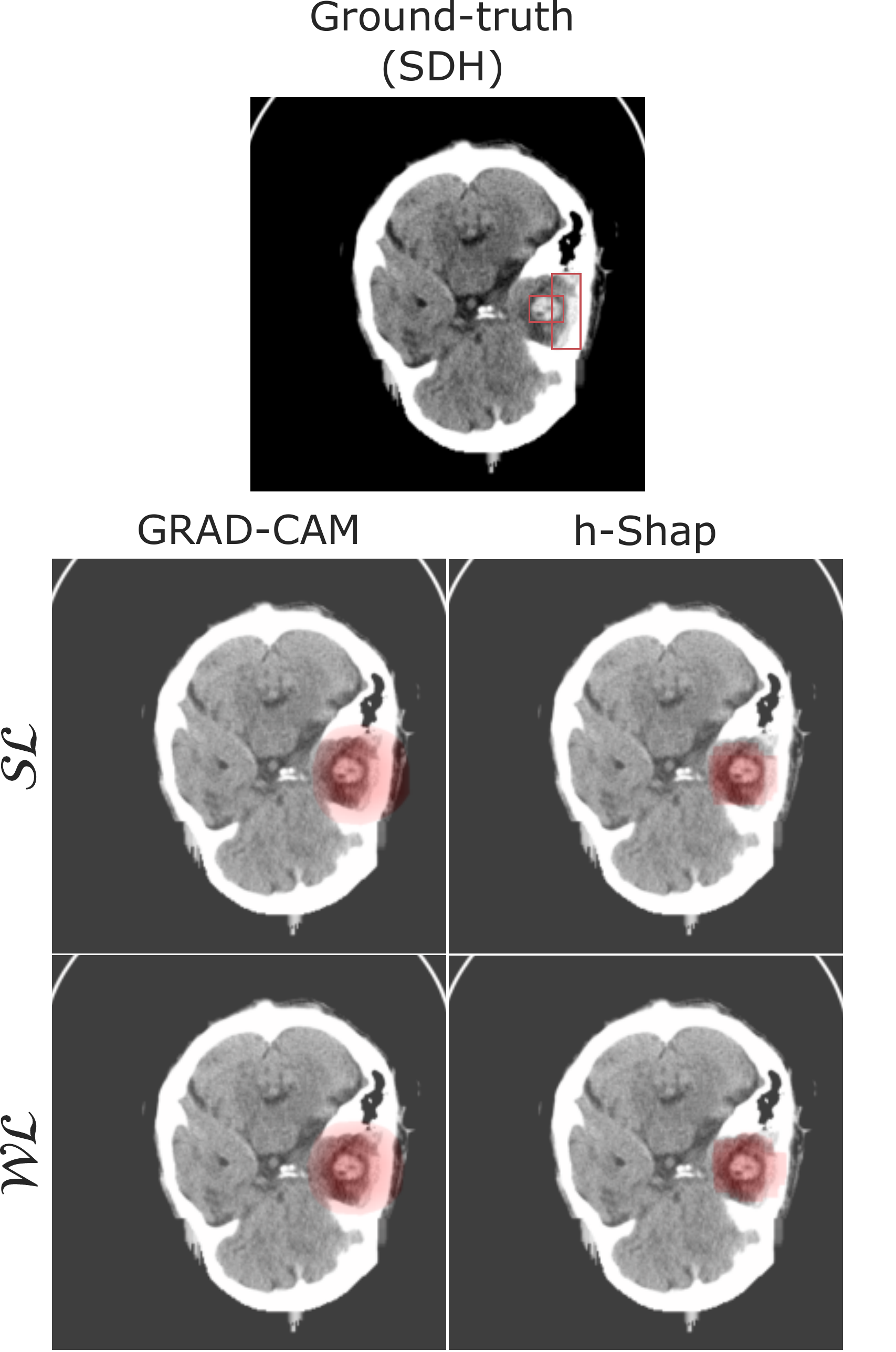}
        \caption{CT-ICH dataset.}
    \end{subfigure}
    \caption{\label{fig:saliency_map_demo}Example saliency maps on some predicted positive images that contain hemorrhage. Saliency maps where obtained by applying GRAD-CAM to the last convolutional layer of both the strong learner ($\SL$) and the weak learner ($\WL$), and by means of h-Shap. All saliency maps are thresholded using Otsu's method to reduce noise.}
\end{figure}

\subsection{MIL provides comparable performance on image-level hemorrhage detection}
\emph{Image-level} hemorrhage detection refers to the localization of the hemorrhage, or its signs, within the selected images (e.g., hyperdense regions, spots, asymmetries). Recall that in this work, we train learners on binary classification tasks either locally (on images for the strong learner $h$), or globally (on examinations for the MIL learner $H$) rather than on a segmentation task. In particular, the RSNA dataset does not provide ground-truth segmentations of the bleeds. In this context, one can attempt to localize the signs of hemorrhage by \textit{explaining} the models' predictions and find their most important features (i.e., pixels) in order to bridge classification with detection. To this end, we use and compare two machine learning explainability methods \cite{burkart2021survey,linardatos2020explainable}: $(i)$ Grad-CAM \cite{selvaraju2017grad}\footnote{Grad-CAM is available at \url{https://github.com/jacobgil/pytorch-grad-cam}.}, which is a saliency method based on sensitivity analysis and very popular in radiology \cite{panwar2020deep,deepika2022deep,chien2022usefulness}, and  $(ii)$ h-Shap \cite{teneggi2022fast}, an efficient hierarchical extension to Shapley-based image explanations that provably retains several of the theoretical benefits of game theoretic explanations (see \cref{sec:image_level_hshap} for details). To our knowledge, this is the first time explanation methods have been used to explain the predictions of a bag-level \cite{amores2013multiple} MIL classifier at the pixel-level \cite{cheplygina2019not}. Even though the saliency maps produced by explanation methods provide a weaker sense of localization, they allow users to interpret a model's prediction and investigate their complex mechanisms.

\begin{figure}[t]
    \centering
    \subcaptionbox{\label{fig:image_level_hem_detect_CQ500_f1}CQ500 dataset.}{\includegraphics[width=0.49\linewidth]{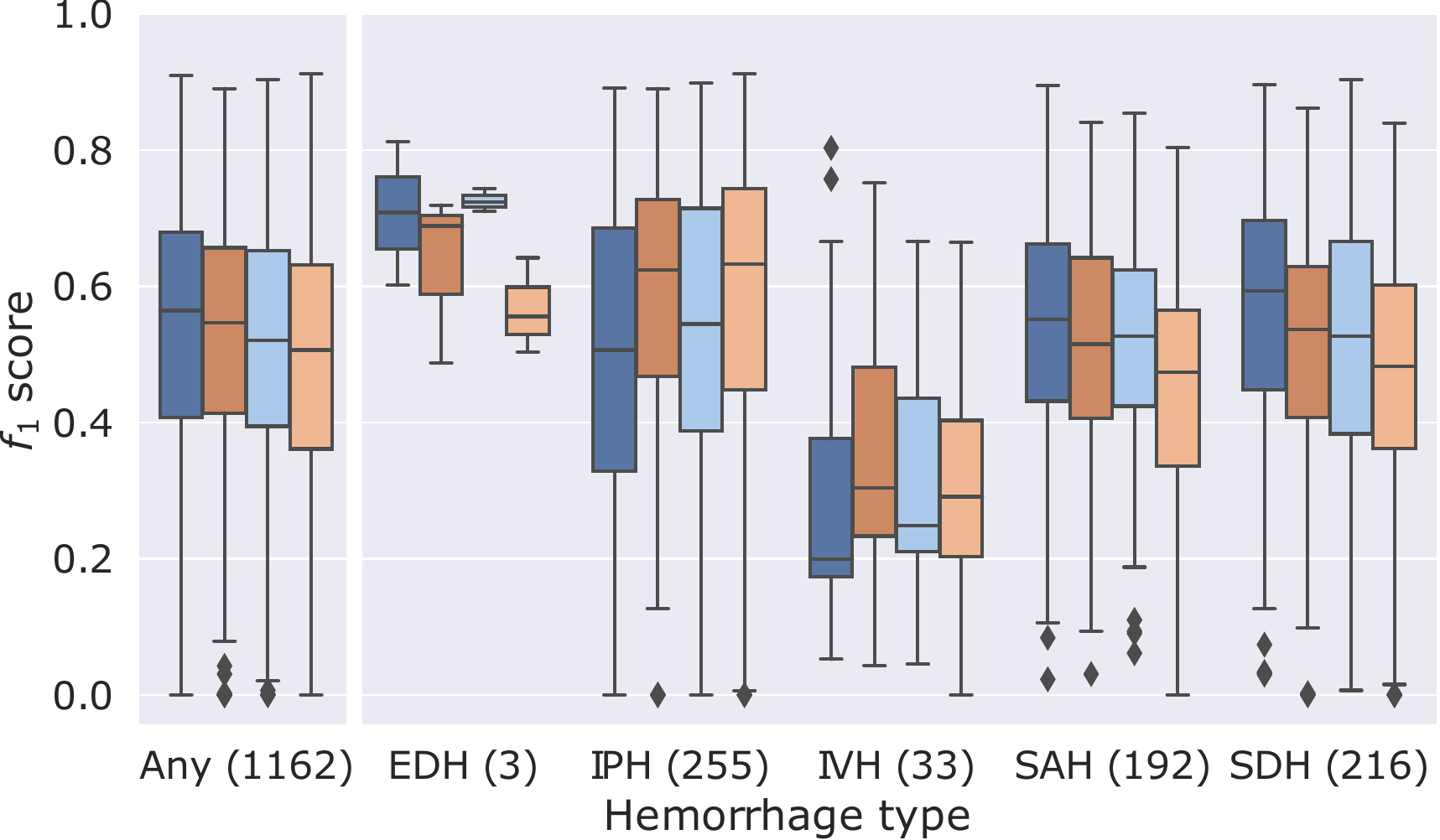}}
    \subcaptionbox{\label{fig:image_level_hem_detect_CT-ICH_f1}CTICH dataset.}{\includegraphics[width=0.49\linewidth]{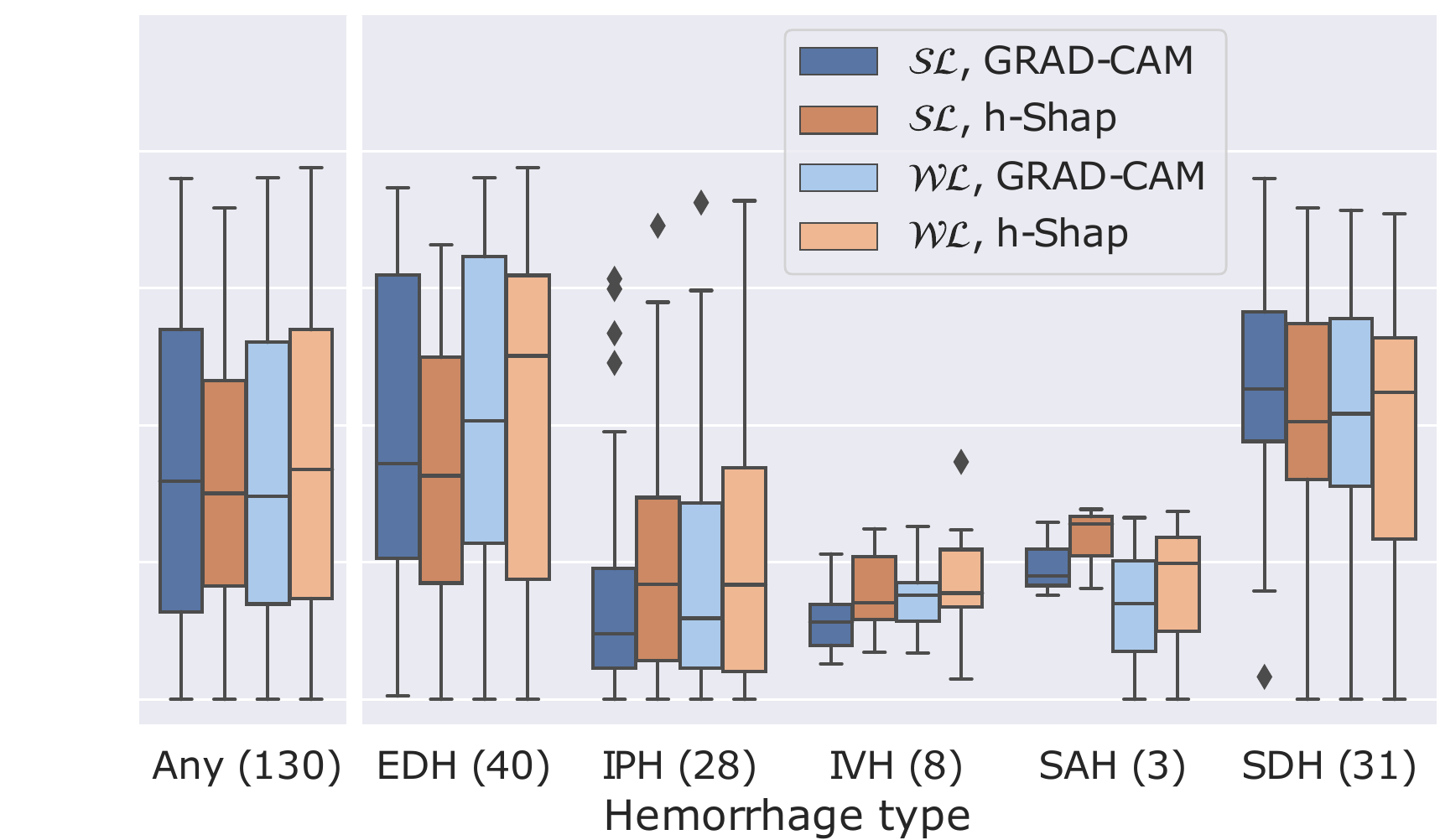}}
    \caption{\label{fig:image_level_all_f1}Image-level hemorrhage detection performance for a strong learner ($\SL$) and weak learner ($\WL$) on the CQ500 and CT-ICH datasets with saliency maps obtained with GRAD-CAM and h-Shap. The $f_1$ scores are computed between the binarized saliency maps and the ground-truth bounding box annotations. For a fair comparison, we show the $f_1$ score distributions of true positive images explained by both models (i.e., 1,162 images for the CQ500 dataset and 130 images for the CT-ICH dataset).}
\end{figure}

\cref{fig:saliency_map_demo} presents an example of saliency maps for both strong and weak learners, using both Grad-CAM and h-Shap, for every type of hemorrhage in the CQ500 and CT-ICH datasets: epidural (EDH), intraparenchymal (IPH), intraventricular (IVH), subarachnoid (SAH), and subdural (SDH). We remark that the CT-ICH dataset provides manual segmentations of the ground-truth lesions, while we use the BHX extension of the CQ500 dataset to obtain ground-truth bounding-boxes for this latter case. \cref{fig:saliency_map_demo} demonstrates that saliency maps produced by either predictor align well with the ground-truth annotations, with no clear advantage of the strongly supervised model. Interestingly, we can also appreciate how the saliency maps concentrate around the target lesions rather than other findings that may correlate well with the presence of ICH in the training set (such as external hematomas due to injury, midline shift effects, or compression of the ventricles).

We further quantitatively evaluate the alignment of the binarized saliency maps with the ground-truth annotations via pixel-level $f_1$ scores. \cref{fig:image_level_all_f1} depicts the distribution of these $f_1$ scores stratified by hemorrhage type for the CQ500 and CT-ICH datasets. For a fair comparison between the strong learner ($\SL$) and the MIL weak learner ($\WL$) we show the distribution of the $f_1$ scores for true positive images that were both predicted by the strong learner to contain signs of hemorrhage \textit{and} selected by the weak learner via Shapley coefficients thresholding. In particular, we compare 1,162 images from the CQ500 dataset and 130 images from the CT-ICH dataset. \cref{fig:image_level_all_f1} confirms that there is no clear advantage of strong supervision for image-level hemorrhage detection. Specifically, there is no combination of learner and explanation method that consistently outperforms all others across all types of hemorrhage and datasets.
Thus, these results suggest that image-level hemorrhage detection can be performed with comparable performance completely without image-level labels.

\subsection{Attention-based MIL significantly reduces the number of labels required}
So far, we investigated the performance of a strong and a weak learner trained on the entire training split of the RSNA dataset---as typically done in these applications \cite{shehab2022machine,burduja2020accurate,gudigar2021automated}. We now study this behavior as a function of the number of labels available to each model during training, $m$. For the strong learner, $m$ thus refers to the number of labeled images, whereas for the weak learner, which only has access to examination-level information, $m$ refers to the numbers of labeled examinations. Note that this quantification is useful because, if the cost of obtaining a label is comparable in both image- and examination-wise cases, this number $m$ reflects an overall cost associated with the annotation of a dataset. Note that in practice, it is a much easier and faster task for an expert radiologist to quickly scroll through the images in an examination and determine whether the whole scan has signs of hemorrhage, rather than having to label all the images in the scan individually. In order to account for the increased variance of the training process with smaller number of samples, we repeat the training process an increasing number of times on random subsets of images or examinations as $m$ decreases (see \cref{sec:training_multiple_times} for details). The obtained models are evaluated on a fixed subset of 1,000 examinations from the validation split of the RSNA dataset.

\begin{figure}[t]
    \centering
    \subcaptionbox{RSNA dataset}{\includegraphics[width=0.3\linewidth]{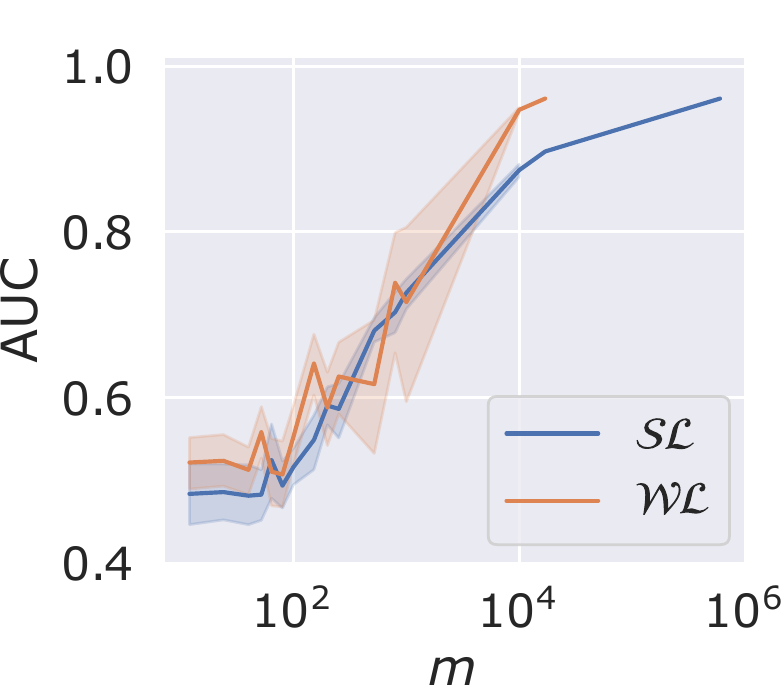}}
    \subcaptionbox{CQ500 dataset}{\includegraphics[width=0.3\linewidth]{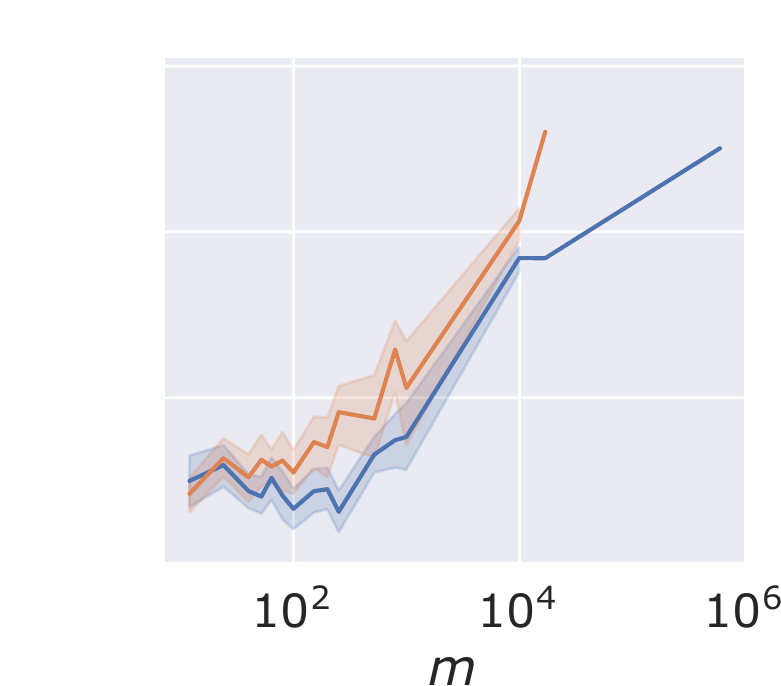}}
    \subcaptionbox{CT-ICH dataset}{\includegraphics[width=0.3\linewidth]{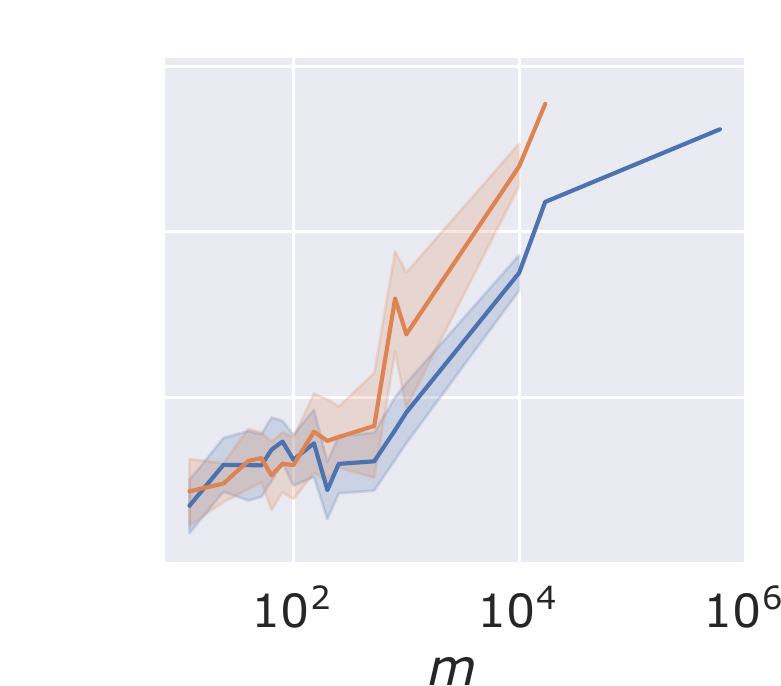}}
    \caption{\label{fig:global_all_auc_label_complexity}Mean performance and 95\% confidence interval of strong ($\SL$) and weak ($\WL$) learners on the examination-level binary classification problem as a function of number of labels $m$. For the RSNA dataset, we validate models on a fixed subset of 1,000 examinations. Note that the points after $m = 1 \times 10^4$ have zero variance because we repeat the training process only once.}
\end{figure}

\cref{fig:global_all_auc_label_complexity} shows the mean AUC's of strong and weak learners, with their 95\% confidence intervals, on the examination-level binary classification problem as a function of the number of labels available to each model during training, $m$. MIL learners show a slight advantage over strong learners on the CQ500 and CT-ICH datasets, while they overlap for the most part with strong learners on the validation split of the RSNA dataset. Furthermore, the performance of MIL learners show a larger variance compared to strong learners. These results suggest that although MIL learners can provide comparable or better performance than strong learners on the examination-level binary classification problem, they might be harder to train. This agrees with intuition that the MIL framework does provide a weaker sense of supervision, and the learners might need to disambiguate the true concept (i.e., ICH) from others that might correlate well with examination-level labels.

\begin{figure}
    \centering
    \includegraphics[width=0.75\linewidth]{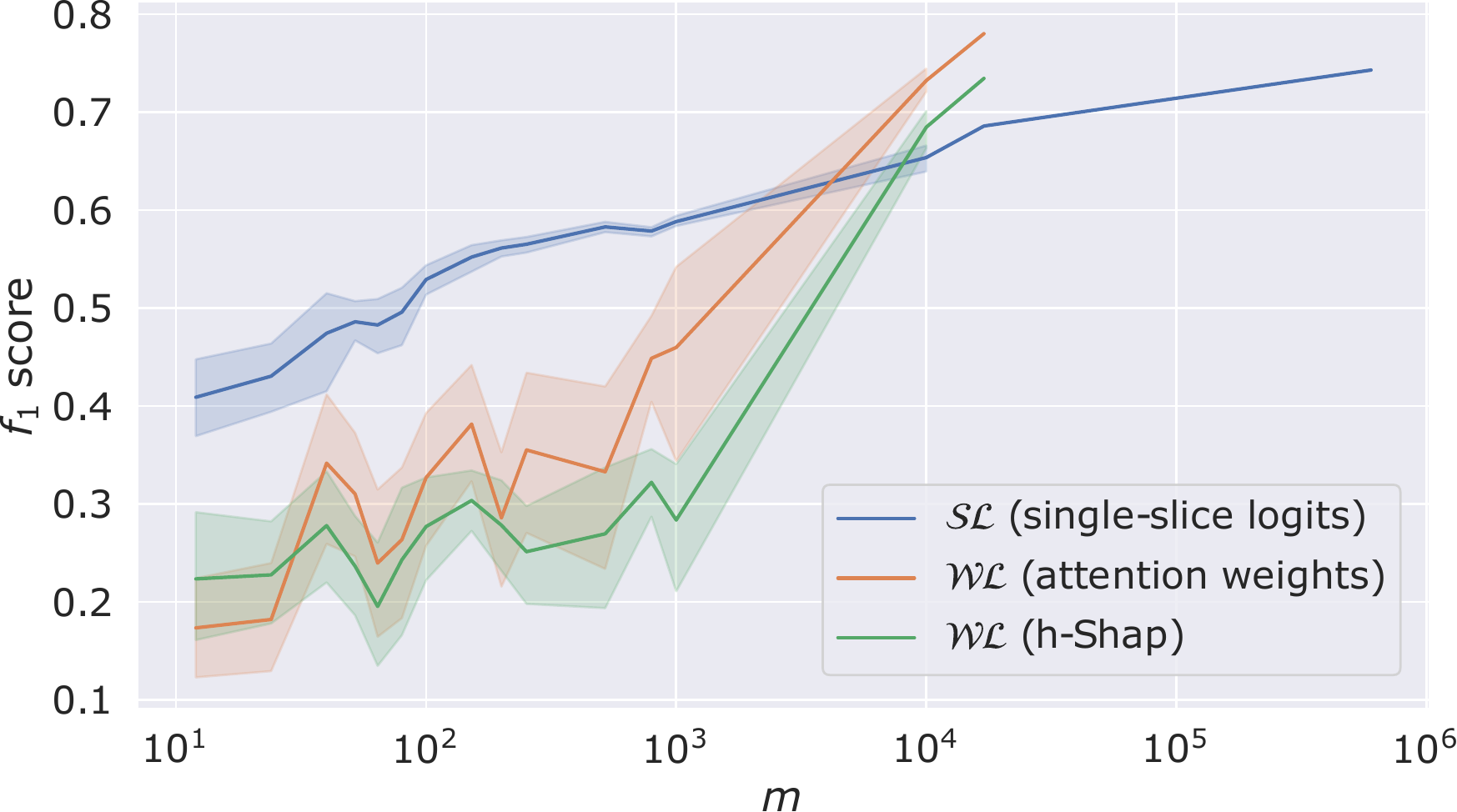}
    \caption{\label{fig:exam_level_RSNA_f1_label_complexity}Mean examination-level hemorrhage detection $f_1$ score as a function of number of labels $m$ on a fixed subset of 1,000 examinations from the validation split of the RSNA dataset.}
\end{figure}

Finally, \cref{fig:exam_level_RSNA_f1_label_complexity} depicts the mean hemorrhage-level detection $f_1$ scores and their 95\% confidence intervals over the validation split of the RSNA dataset as a function of the numbers of labels available. Confidence intervals are computed across repetitions of the training process with the same number of labels, thus capturing the variance of the training process. Since we train only one model for $m > 10^4$ labels, the variance vanishes. We see that for $m \leq 10^4$, strong supervision does in fact provide a significant advantage over weak supervision. However, MIL learners quickly outperform strongly supervised ones for $m \gtrsim 10^4$. Importantly, these results confirm that attention-based models trained on examination-level binary labels can provide comparable performance to traditional classifiers trained on image-level labels while requiring $\approx 35$-times fewer labels. Note that the curves for the weak learners interrupt after $10^4$ labels because they reach the limit of the training data size---the training split of the RSNA dataset contains $\approx 17 \times 10^4$ labeled examinations, whereas there are about $35$ times more image labels.

\section{Discussion}
In this study, we compared the performance of predictive models for ICH detection in Head CT scans trained with strong supervision (i.e., having one label per image within an examination) or weak supervision (i.e., using a single label for each entire examination). The methodology is based on recent Multiple Instance Learning (MIL) approaches via attention mechanisms \cite{vaswani2017attention,ilse2018attention}, which strictly generalizes predictors that are trained with strong supervision. This framework enabled the use of models that have the same architecture and main components in either setting, making these comparisons precise and fair. We found that weakly supervised models had comparable performance to strongly supervised models, despite using approximately 35 times fewer labels. On one external dataset, the weakly supervised models actually had significantly higher performance, suggesting better generalizability. Importantly, weakly supervised models also had comparable ability to localize ICH on image- and pixel-levels. Altogether, these findings indicate that image-level annotations are not necessary to train high-performing and explainable DL models for diagnosis of ICH on head CT.

Our first result demonstrated that strong supervision is not at all necessary for weak, or global, prediction tasks, as long as sufficient data is available. More precisely, \cref{fig:global_all_roc} demonstrates that whether a predictor is trained on image-wise or examination-wise labels, they obtain virtually the same AUC in the task of predicting the presence or absence of ICH. In one of the studied datasets (CT-ICH \cite{hssayeni2020computed}), there is in fact a slight advantage to the latter, suggesting better generalizability to different clinical populations. This is not surprising, as the weak learner is precisely trained on the task of detecting hemorrhage at the examination level. Nevertheless, this generalizability advantage is important, given the well-documented drops in performance of DL models for medical imaging diagnosis on external test sets \cite{zech2018generalization}, which threaten the safe deployment of DL models in real-world clinical practice.

Interestingly, our results further demonstrate that these observations hold even for the case of examination-level hemorrhage detection, i.e. the task of finding the images within each examination where signs of hemorrhage are present. More precisely, even though the MIL model was only trained on global examination-wise labels, one can make predictions about each of the constituent images by either studying their attention weights, or by employing game-theoretic tools like the Shapley value \cite{Shapley1953,lundberg2017unified}. In either case, the ability to detect the positive images within positive examination is comparable to the performance of a model trained with strong supervision, i.e. with a label per image (\cref{fig:exam_level_CT-ICH_TPR}). We emphasize that the ability to identify examination-level hemorrhage is of critical practical importance for radiologists' workflows. In current triage use cases of DL models in radiology, cases with potentially actionable findings, such as ICH, are flagged by DL models for radiologist review, after which a radiologist must confirm whether they agree or disagree with the diagnosis. Having image ``flags'' beyond the mere presence or absence of hemorrhage that show which specific images have a hemorrhage prediction are critical for: $(i)$ allowing a radiologist to expeditiously confirm presence or absence of hemorrhage, and $(ii)$ building trust with radiologists and other physician end-users, who have been shown to be less trusting of diagnostic results generated by automated systems in medical imaging compared to those provided by human experts \cite{Gaube2021DoAsAiSay}.

When further analyzing the resulting models with both a popular saliency method (Grad-CAM \cite{selvaraju2017grad}) as well as newer approaches to interpretability with theoretical guarantees (h-Shap \cite{teneggi2022fast}), both models indicate having captured the same semantic concepts that constitute ICH in head CT (\cref{fig:saliency_map_demo}). Indeed, our qualitative and quantitative comparisons of these saliency maps indicate that the ability to find the corresponding hemorrhages within each of the images is virtually the same, and only mild differences exist once stratified per ICH type (\cref{fig:image_level_all_f1}). To this point, we remark that to verify whether a model did learn the desired concept (i.e., ICH) instead of other spurious correlations in the training data is especially important in medical imaging. As modern machine learning models continue becoming increasingly complex, gaining insights on their decision making process is fundamental for a responsible use in real-world scenarios. As discussed above, building trust with physician end-users is paramount, and providing pixel-level explanations for specific disease identification will be important towards this goal. Furthermore, medical institutions may be required by certain laws to provide explanations of what lead an automatic systems to recommend a certain treatment or to provide a specific diagnosis \cite{goodman2017european}.

Our work also has limitations. First, we evaluated only a single diagnostic use case of ICH detection on CT scans of the head, albeit with multiple datasets from different clinical populations. However, our approach is applicable to any other medical imaging use case that utilizes cross-sectional imaging, including diagnosis of disease on CT of other body parts, as well as on other imaging modalities, such as MRI. Future studies will apply our approach to other use cases to validate its generalizability in other diagnostic scenarios and imaging modalities. Second, while this study demonstrated that indeed the examination-level annotations suffice for ICH detection in CT once enough training data is available, \textit{some} image-level annotations were needed to validate our methodology. In future extensions to other diagnostic tasks or imaging modalities, this minimal amount of locally annotated data will also be necessary for validation purposes. This number of local annotations is very small, however: in this work we employed 1,000 examinations of the validation split of the RSNA dataset to this end, requiring about $35 \times 10^3$ image-level labels. This represents less than 6\% of the number of image-level labels needed to train an alternative strongly supervised model. Third, the weakly supervised method currently only evaluates medical imaging data; given the potential improvements in imaging diagnoses using multimodal AI models \cite{acosta2022multimodal} incorporating multiple types of medical data (e.g., imaging, clinical symptoms, laboratory values), developing weakly supervised DL models that can incorporate multiple data types is an important topic for future study. Finally, although Convolutional Neural Networks (CNNs)---such as the models used in this work---remain the most popular deep learning architecture in medical imaging, it remains important to investigate whether these results extend to other parametrization of the predictors and architecture choices, for example to Vision Transformers (ViTs) \cite{dosovitskiy2020image,khan2022transformers}---which are rapidly gaining popularity in the field.

In summary, our results indicate that training DL models with weak or strong supervision provides comparable performance for the tasks of ICH detection in head CT across three different levels of granularity: $(i)$ global binary prediction, $(ii)$ examination-level detection, and $(iii)$ image-level detection. Our last results explore these points further by studying the performance of strong and weak learners on the global binary classification problem, as well as on examination-level hemorrhage detection as a function of the number of labels available during training. These results indicate that, indeed, weakly supervised learning enables a significant reduction in the need for annotations: once the number of labels provided is large enough ($m \gtrsim 5 \times 10^3$) weakly supervised models achieve comparable performance to strongly supervised models at a fraction of the provided labels. However, for the strongly supervised predictor, these labels represent the number of labeled images, whereas the for the MIL predictors, $m$ represents only global information of the entire examination---which can be easily collected, e.g. from clinical reports. This approach could apply to other 3D cross-sectional imaging tasks, such as MRI diagnosis, potentially saving thousands of hours of annotation labor by radiologists \cite{flanders2020construction}, thereby alleviating the biggest bottleneck in developing high-performing DL models for medical imaging diagnosis.

\section{Methods}
\label{sec:methods}

\subsection{Learning paradigms}
\label{sec:learning_paradigms}

\subsubsection{Supervised learning}
In supervised learning settings, given input and output domains $\X$ and $\Y$, one is interested in predicting a response $y \in \Y$ on an input $x \in \X$ with a predictor $h:~\X \to \Y$. Given a loss function $\ell(y, h(x))$ that penalizes the dissimilarity between the true label $y$ and the predicted label $h(x)$, we search for a predictor $h$ with low risk over a suitable family of predictors (e.g., Convolutional Neural Networks). This search is usually carried out by minimizing the empirical loss over a training set $\{(x_i, y_i)\}_{i=1}^{n_s}$ such that
\begin{equation}
    h = \argmin_{h'} \frac{1}{n_s}\sum_{i=1}^{n_s} \ell(y_i, h'(x_i)).
\end{equation}

\subsubsection{Multiple-Instance Learning (MIL)}
Multiple Instance Learning (MIL) \cite{dietterich1997solving,maron1997framework,weidmann2003two} generalizes the supervised learning framework to \emph{bags} of inputs. Formally, recall that $\X$ and $\Y$ are input and output domains, and let $X = (x^{(1)}, x^{(2)}, \dots, x^{(r)}) \in \X^r$, $r \in \N$ indicate a bag with $r$ \emph{instances}. Furthermore, the MIL paradigm assumes that the bag-level response $Y \in \Y$ is a known function of the instance-level responses $y^{(1)}, y^{(2)}, \dots, y^{(r)}$, which can encompass a wide variety of choices \cite{blum1998note,auer1998approximating,andrews2002support,sabato2009homogeneous,sabato2012multi,Wang_MIL_2022}. 

In this work, we focus on MIL binary classification, such that
\begin{equation}
    \label{eq:binary_mil_assumption}
    Y = \texttt{OR}(y^{(1)}, y^{(2)}, \dots, y^{(r)}),
\end{equation}
and we search for a \emph{bag-level} classifier $H:~\X^r \to \Y$ with low risk over a suitable class of predictors. Similarly to the supervised learning paradigm, given a loss function $\ell(Y, H(X))$ that penalizes wrong predictions, $H$ is found by optimization of the empirical loss over a training set $\{(X_i, Y_i)\}_{i=1}^{n_w}$ of labeled bags, such that
\begin{equation}
    H = \argmin_{H'} \frac{1}{n_w} \sum_{i=1}^{n_w} \ell(Y_i, H'(X_i)).
\end{equation}
Importantly, we remark that an MIL learner does not have access to the underlying instance-level labels. Finally, note that:
\begin{itemize}
    \item The examination-level binary classification problem satisfies the MIL assumption in \cref{eq:binary_mil_assumption}, as the global label of an examination is positive as soon as it contains at least one image with signs of hemorrhage (i.e., a positive image). Equivalently,
    \item The local image-level labels can also be phrased as an instance of \cref{eq:binary_mil_assumption}. In particular, an image should be labeled positively as soon as it contains signs of hemorrhage.
\end{itemize}

\subsection{Model architecture details}
\label{sec:model_architecture_details}
\subsubsection{Strong learner}
The strong learner $h$ is the composition of a feature extractor $f$ with a binary classifier $g$ implemented by a fully connected layer with sigmoid activation. In this work, $f$ is a ResNet18 \cite{he2016deep} pretrained on ImageNet \cite{deng2009imagenet} that encodes an input image of size $512 \times 512$ pixels into a vector of size 256, as illustrated in \cref{fig:sl_architecture}.

\subsubsection{Weak learner}
In addition to the same feature extractor $f$ and final classifier $g$ employed for the strong learner, the weak learner $H$ comprises a two-layer attention mechanism $a$ as proposed in \cite{ilse2018attention} (see \cref{fig:wl_architecture}). For an input examination with $r$ images, the attention mechanism combines the $r$ image-level feature vectors into a single examination-level feature vector which can be expressed as a convex combination of the image-level feature vectors. In this work---differently from the original work in \cite{ilse2018attention}---we use the \emph{sparsemax} activation function \cite{martins2016softmax,peters2019sparse,correia19adaptively} rather then the softmax function to favor sparse attention weights.\footnote{The entmax package is available at: \url{https://github.com/deep-spin/entmax}.}

\subsection{Data preprocessing}
\label{sec:data_preprocessing}
The images in the three datasets used in this work were annotated by expert neuroradiologists of varying degree of expertise with the type(s) of hemorrhage present in the image. We group the original classes into \texttt{`normal'} (i.e., label 0, no type of hemorrhage) and \texttt{`with hemorrhage'} (i.e., label 1, any type of hemorrhage). Images are provided in DICOM and NIfTI format, so we:
\begin{enumerate}
    \item Convert them to Hounsfield Units (HUs) \cite{buzug2011computed}, then
    \item Window them using the standard brain window setting, i.e. WL = 40 and WW = 80 \cite{turner2011ct}, and finally
    \item Normalize them with min-max normalization.
\end{enumerate}
This way, pixel intensities represent the same HU value (and hence, tissue) across all datasets.

\subsection{Training procedures}
\label{sec:training_procedures}
Experiments were performed on Nvidia Quadro RTX 5000 GPU's on a private cluster and on the Azure Machine Learning (ML) platform \cite{barnes2015azure} via Microsoft Research's Project InnerEye open-source software tools (\url{https://aka.ms/InnerEyeOSS}).\footnote{Project InnerEye is available at: \url{https://github.com/microsoft/InnerEye-DeepLearning}.}

To account for the high label imbalance in the training split of RSNA dataset and for the gap in difficulty between the prediction of the presence of hemorrhage compared to predicting its absence, models were trained using \emph{focal loss} \cite{lin2017focal}---a variation of binary cross-entropy loss. All models were trained for 15 epochs with a learning rate decay of 0.3 every 3 epochs. We chose the best performing model across epochs according to validation accuracy.

\subsubsection{Image-level augmentation}
We use TorchIO's \cite{perez2021torchio} library of spatial and intensity transformations.\footnote{TorchIO is available at \url{https://github.com/fepegar/torchio}.} Specifically, every image is augmented independently via random flips, affine transformations, deformations and rotations, and one out of addition of random noise, random bias field, random anisotropy, random gamma transformation, or random ghosting artifacts.

\subsubsection{Examination-level augmentation}
We randomly sample without replacement between 10 and $r$ images within the same examination. This sub-sampling augmentation strategy does not rely on image-level labels and it can be used in weakly supervised scenarios where only examination-level labels are available. Intuitively, sampling at least 10 images controls the probability of flipping a positive examination to a negative one. That is, sampling a subset of all negative images from a positive examination would result in a wrong label (i.e., the subset would be labeled positively even if it did not contain any positive images). Although we cannot completely rule out this event without knowing local labels, we can reduce its probability to a tolerable level for the weak learner. Formally, given a positive examination $X$ of length $r$ with $K$ positive images, sample a random subset $S$ of images without replacement. Denote $Y_S$ the true global label of the subset, and note that $p_{\text{flip}} = \P[Y_S = 0 \mid Y = 1]$ is a decreasing function of the size of the subset $S$ and it follows a hypergeometric distribution. In this work, we estimate $p_{\text{flip}}$ over the training split of the RSNA dataset, and obtain $p_{\text{flip}} \leq 4 \times 10^{-3}$. We remark that to estimate $p_{\text{flip}}$ we used the image-level labels provided in the training split of the RSNA dataset. In practical scenarios this information can easily be replaced by prior knowledge of expert radiologists about the problem at hand.

\subsubsection{Training strong learners}
Strong learners were trained using Adam optimizer \cite{kingma2014adam} with learning rate of $1 \times 10^{-5}$, weight decay of $1 \times 10^{-7}$, and batch size of 64. During training, we add a dropout layer with $p = 0.50$ between the encoder $f$ and the binary classifier $g$.

\subsubsection{Training weak learners}
Weak learners were trained using Stochastic Gradient Descent (SGD) with momentum equal to $0.9$ \cite{sutskever2013importance}, learning rate of $1 \times 10^{-3}$, weight decay of $1 \times 10^{-4}$, and batch size of 1. We remark that the choice of batch size equal to 1 comes both from memory limitations and gradient propagation imbalances through the attention mechanism for volumes with different numbers of images. During training, we add both a dropout layer with $p = 0.50$ between the encoder $f$ and the binary classifier $g$ and a dropout layer with $p = 0.25$ after the first layer of the attention mechanism.

\subsection{Explaining model predictions with h-Shap}
We use h-Shap \cite{teneggi2022fast}, a Shapley-based explanation method with provable runtime and accuracy for problems that satisfy the binary MIL assumption in \cref{eq:binary_mil_assumption} to select the positive images in an examination, and to highlight signs of hemorrhage within the selected images.

\subsubsection{Examination-level hemorrhage detection}
\label{sec:exam_level_hshap}
We extend the original implementation of h-Shap to explain the examination-level prediction of a weak learner. Since the global binary label satisfies \cref{eq:binary_mil_assumption}, one can explore a binary tree of the input examination and hierarchically compute the exact Shapley coefficient of every image in the examination \cite[see][Theorem~3.4]{teneggi2022fast}. The symmetry axiom of the Shapley value \cite{Shapley1953} implies that the positive images in an examination should receive the same coefficient. Thus, one can use an importance threshold $t = 1/r$ and select those images whose Shapley values are $\geq t$. We remark that---as recently noted by others \cite{covert2022learning,jain2022missingness}---explaining predictions on sets with the Shapley value is particularly attractive because it does not require to sample an uninformative baseline to mask features \cite{lundberg2017unified}. In fact, the weak learner can predict on sequences of arbitrary length and it is permutation invariant \cite{zaheer2017deep}, hence one can simply remove images from an examination without having to replace them.

\subsubsection{Image-level hemorrhage detection}
\label{sec:image_level_hshap}
Bleeds can present complex and irregular shapes. However, h-Shap explores fixed quad-trees of the input image. Thus, we extend the original implementation with standard ideas of cycle spinning \cite{coifman1995translation}. Denote $s$ the minimal feature size in h-Shap (i.e., the size of the smallest leaf explored by the algorithm), and let $\bm{\rho} = \{\rho_i\}_{i=1}^{n_{\rho}}$ be $n_{\rho}$ equally spaced radii between 0 and $s$, and let $\bm{\alpha} = \{\alpha_i\}_{i=1}^{n_{\alpha}}$ be $n_{\alpha}$ equally spaced angles between 0 and $2\pi$. Then, we average the saliency maps obtained by cycle spinning the original partition by the vector $(\rho \cos(\alpha), \rho \sin(\alpha))$, $\rho \in \bm{\rho},~\alpha \in \bm{\alpha}$. Finally, we note that we use the unconditional expectation over the training split of the RSNA dataset to mask features, which is a valid choice in MIL binary classification problems \cite{teneggi2022fast}. In this work, we use h-Shap with an absolute importance tolerance $\tau$ equal to 0 (i.e. h-Shap explores all partitions with a positive Shapley coefficient), minimal feature size $s = 64$, number of radii $n_{\rho} = 3$, and number of angles $n_{\alpha} = 12$.

\subsection{Comparing strong and weak learners on examination-level hemorrhage detection}
\label{sec:exam_level_detection_comparison}
In this section we expand on the methodology and choice of parameters for comparing strong and weak learners on examination-level hemorrhage detection. All choices were made to provide a fair comparison between strongly supervised and weakly supervised models.

\subsubsection{Choosing the classification threshold}
Examination-level hemorrhage detection is performed only for predicted positive examinations. Recall that both strong and weak learners are real-valued functions on the unit interval $[0, 1]$. Thus, a threshold $t \in [0, 1]$ (e.g., $0.5$) is required to binarize their predictions. The choice of $t$ induces a False Positive Rate (FPR) and a True Positive Rate (TPR) on images (for a strong learner) or on examinations (for a weak learner). In this work, we use Youden's $J$ statistic \cite{youden1950index,perkins2006inconsistency,habibzadeh2016determining} to find the threshold $t^*$ that maximizes the difference of TPR and FPR, i.e. $J = \text{TPR} - \text{FPR}$. Then:
\begin{itemize}
    \item For a strong learner $h$, we choose the threshold $t^*_s$ that maximizes $J$ on the image-level labels, and,
    \item For a weak learner $H$, we choose the threshold $t^*_w$ that maximizes $J$ on the examination-level labels.
\end{itemize}
We remark that there exist other methods to choose the threshold $t^*$, and the main results discussed in this work do not strongly depend on this choice. For completeness, \cref{fig:exam_level_all_TPR_d,fig:exam_level_RSNA_f1_d_label_complexity} show the equivalent of \cref{fig:exam_level_all_TPR,fig:exam_level_RSNA_f1_label_complexity} where instead of maximizing Youden's $J$, $t^*$ is chosen to minimize the distance to the $(0,1)$ point (perfect classification), which can be written as $d = \sqrt{\text{FPR}^2 + \left(1 - \text{TPR}\right)^2}$ and is also common in the literature \cite{perkins2006inconsistency,habibzadeh2016determining}. 

\subsubsection{Choosing the best minimal sequence length}
To reduce the false positive rate in the predicted hemorrhage sequences, we fine-tune the minimal number of consecutive positive images that have to be selected by each method in order for a series of consecutive selected images to be considered a candidate hemorrhage sequence. We set this length to 4 for strong learner, to 2 for weak learners when using attention weights to select images, and to 3 for weak learners when using Shapley values, guaranteeing the best performance for each method. \cref{fig:exam_level_hem_detection_RSNA_min_seq_length} depicts the examination-level $f_1$ score as a function of this minimal sequence length on the validation split of the RSNA dataset with both Youden's $J$ and distance to the $(0,1)$ point, which motivate these choices.

\subsubsection{Computing the examination-level $f_1$ score}
Denote $T = \{T_1, T_2, \dots, T_n\}$ the set of true hemorrhage sequences (i.e., non-overlapping series of consecutive positive images) in a positive examination. That is, $T_i$ contains the indices of the images in the $i^{\text{th}}$ hemorrhage sequence. Let $S = \{s_1, s_2, \dots, s_r\}$ be the local estimator used to select positive images depending on the type of learner: single-image logits for a strong learner, and either attention weights or Shapley values for a weak learner. Denote $P = \{P_1, P_2, \dots, P_n\}$ the predicted hemorrhage sequences by the learner. We define the True Positive (TP), False Positive (FP), and Predicted Positive (PP) sequences as
\begin{gather}
    \TP \coloneqq \#\{i \in [n]:~\exists P_j \in P:~(\argmax_{k \in P_j}~s_k) \in T_i\}\\
    \FP \coloneqq \#\{j \in [m]:~\nexists T_i \in T:~(\argmax_{k \in P_j}~s_k) \in T_i\}\\
    \PP = \TP + \PP.
\end{gather}
Put into words, for every true hemorrhage sequence $T_i \in T$, we count one true positive prediction if there exists a predicted hemorrhage sequence in $P_j \in P$ such that the image with the larges estimator value within $P_j$ is contained in $T_i$. Note that this definition of TP does not double count predicted sequences that may correspond to the same true one, and using the $\argmax$ avoids the trivial case where a model may select all images, or a few very long sequences that could include multiple true ones. Similarly, we count one false positive prediction for every predicted sequence $P_j$ for which there does not exists a corresponding true one. The $f_1$ score is then defined as the harmonic mean of precision and recall, i.e.
\begin{gather}
    \text{precision} = \frac{\TP}{\PP},\quad \text{recall} = \frac{TP}{\lvert T \rvert}\\
    f_1 = 2 \cdot \frac{\text{precision} \cdot \text{recall}}{\text{precision} + \text{recall}}.
\end{gather}
We note that this procedure reflects how a machine learning model could be deployed in a clinical setting to detect hemorrhage sequences.

\subsection{Training multiple times on the same number of labels}
\label{sec:training_multiple_times}
When training models on a fixed number of labels $m$, we randomly sample without replacement a subset of the original training split of the RSNA dataset (of images for the strong learners, and of examinations for the weak learners) that maintains the label proportions of the original dataset. In particular, we use 15 distinct values of $m$: $\bm{m} = [12,\ 24,\ 40,\ 52,\ 64,\ 80,\ 100,\ 152,\ 200,\ 252,\allowbreak\ 520,\ 796,\ 10^3,\ 10 \times 10^3,\ 17 \times 10^3]$. For each choice of $m$, we train a decreasing number of models to account for the variance in the training process. In particular, we repeat the training process 20 times when $m \leq 252$; 15 times for $m = 520$; 10 times for $m = 796,\ 10^3$; 6 times for $m = 10 \times 10^3$; and 1 time for $m \geq 17 \times 10^3$. Finally, instead of training models for a fixed number of epochs, we set a \emph{patience} parameter such that training is terminated if the validation accuracy of a model does not increase for more than 3 consecutive epochs.

\appendix
\newpage
\clearpage
\section{Figures}
\renewcommand\thefigure{\thesection.\arabic{figure}}
\setcounter{figure}{0}

\begin{figure}[h]
    \centering
    \subcaptionbox{\label{fig:exam_level_hem_detection_RSNA_youden_min_seq_length}Youden's $J$}{\includegraphics[width=0.48\linewidth]{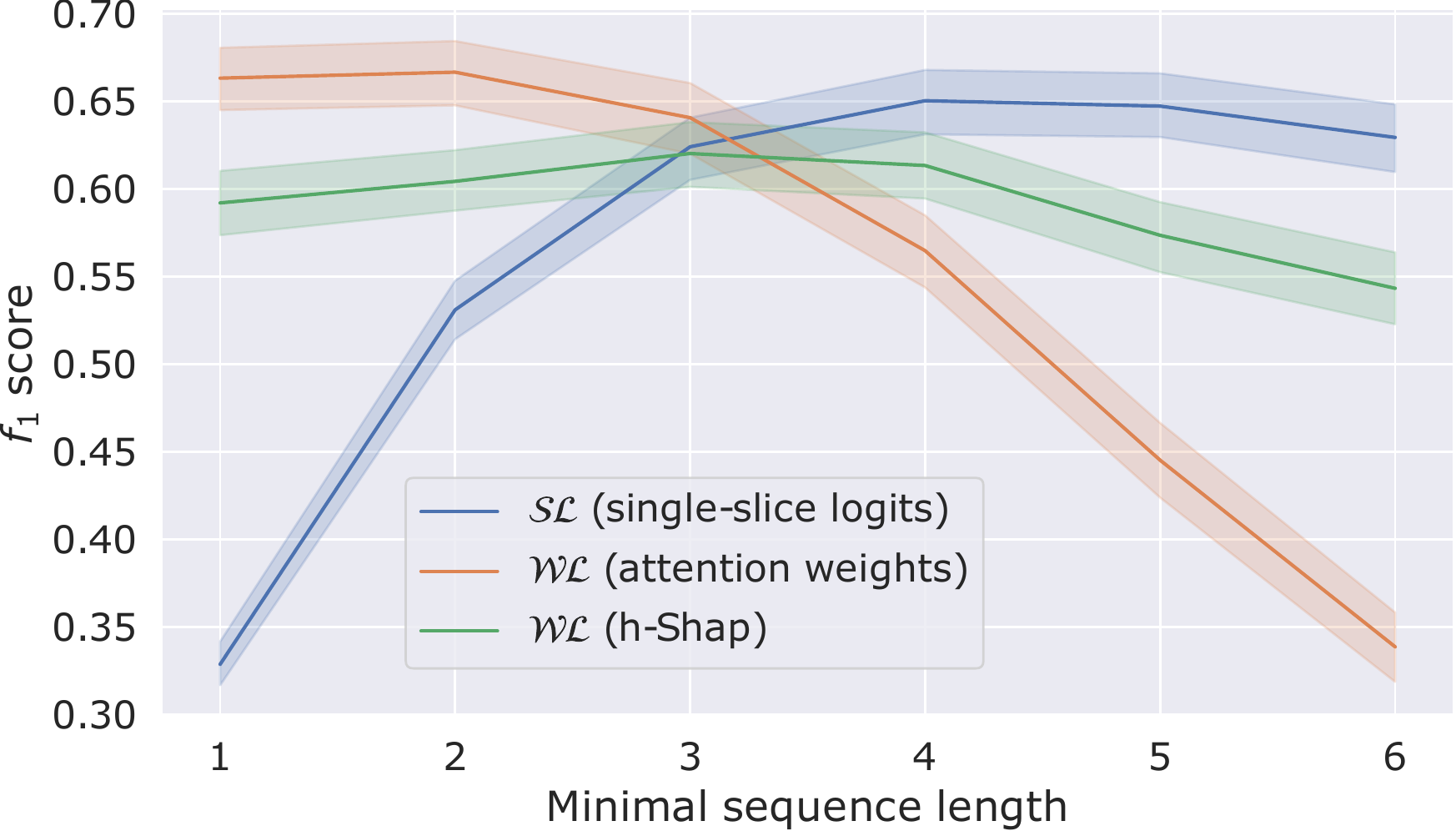}}
    \hfill
    \subcaptionbox{\label{fig:exam_level_hem_detection_RSNA_d_min_seq_length}Distance to $(0,1)$}
    {\includegraphics[width=0.48\linewidth]{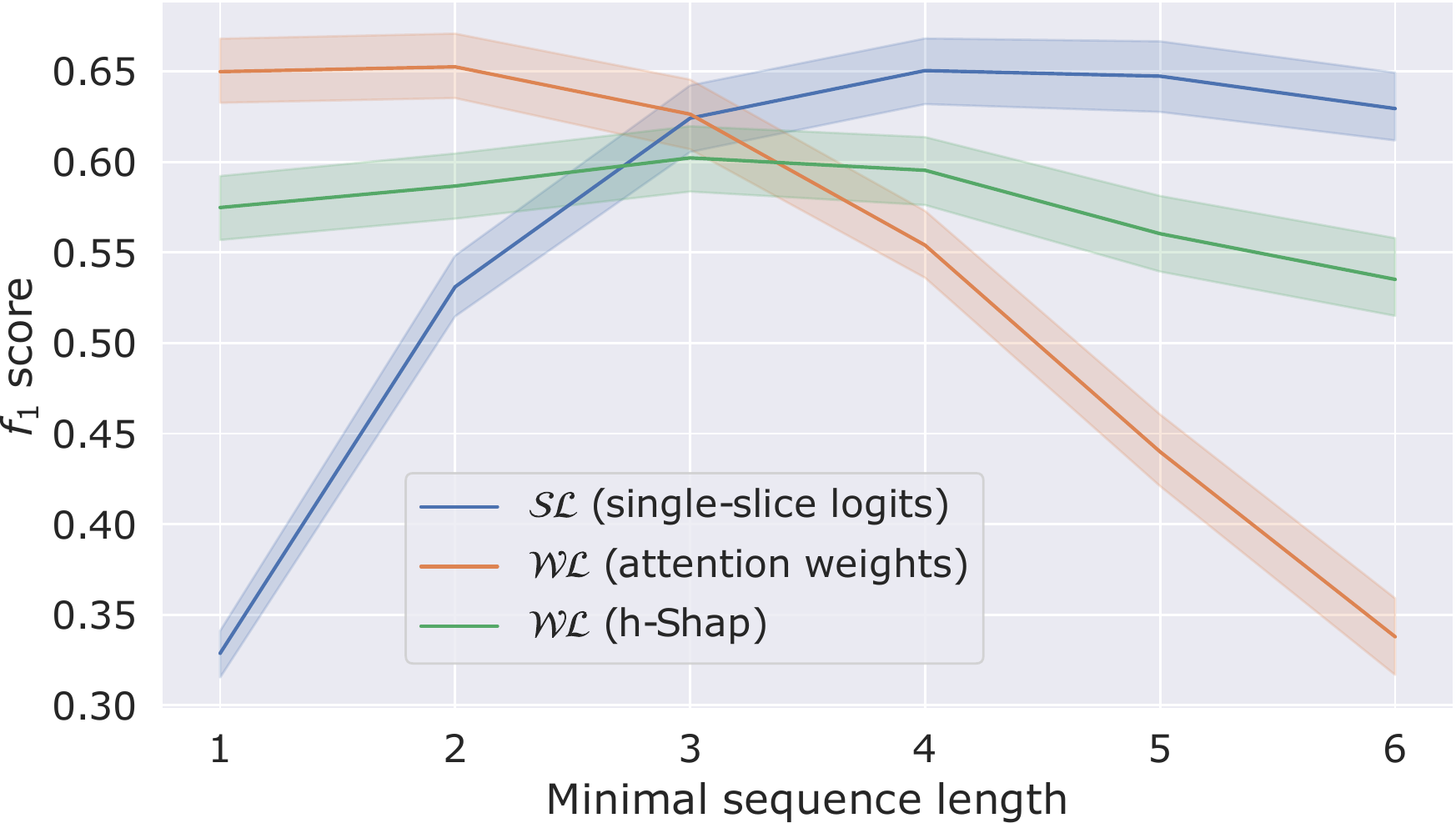}}
    \captionsetup{subrefformat=parens}
    \caption{\label{fig:exam_level_hem_detection_RSNA_min_seq_length}Examination-level $f_1$ score as a function of minimal sequence length for a strong learner ($\SL$) and a weak learner ($\WL$) on the validation split of the RSNA dataset. \subref{fig:exam_level_hem_detection_RSNA_youden_min_seq_length} Results with Youden's $J$ statistic. \subref{fig:exam_level_hem_detection_RSNA_d_min_seq_length} Results with distance to $(0,1)$ point. Note that the best minimal sequence length does not depend on the threshold $t^*$.}
\end{figure}

\begin{figure}[h]
    \centering
    \subcaptionbox{RSNA dataset}{\includegraphics[width=0.47\linewidth]{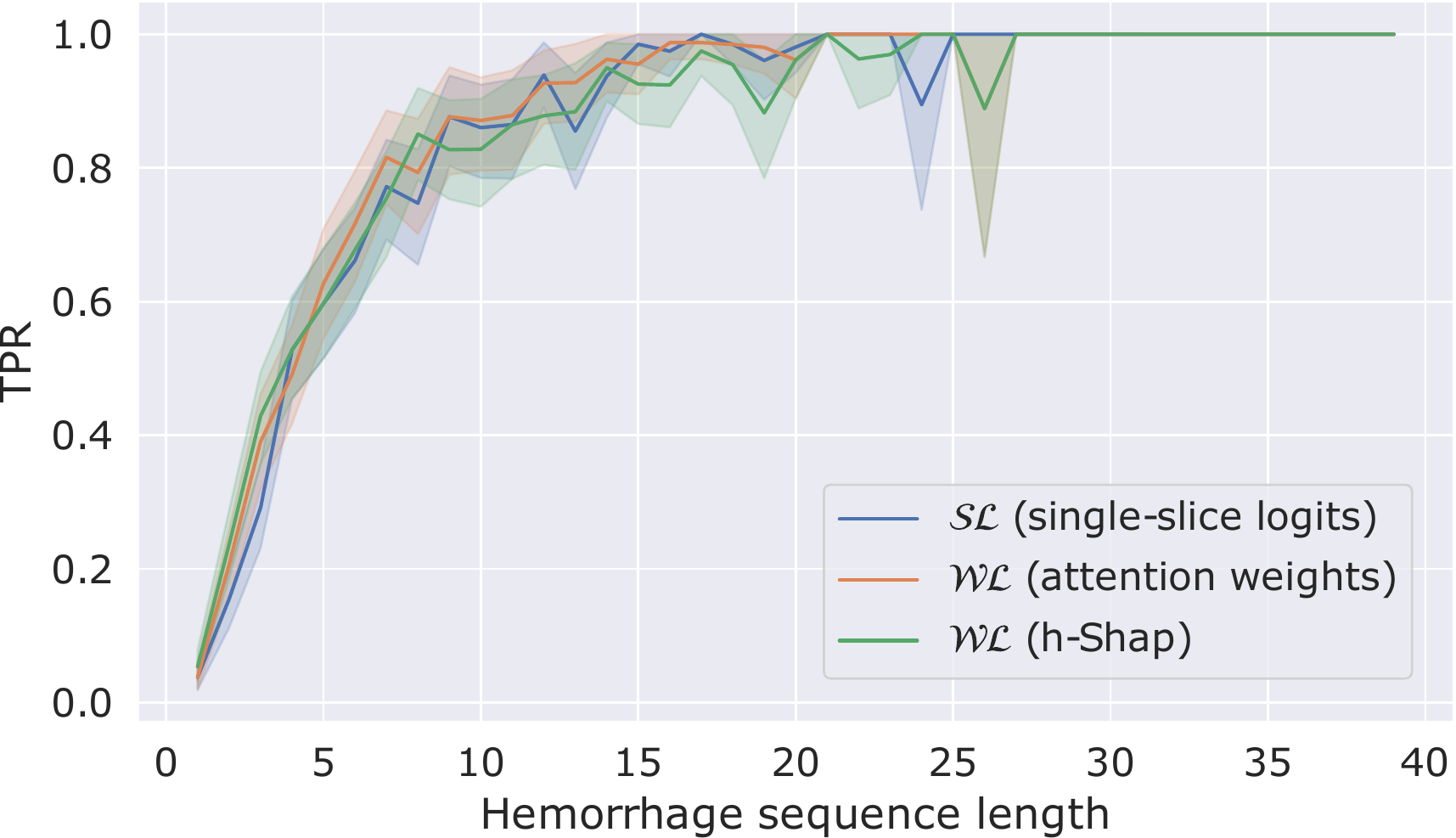}}
    \hfill
    \subcaptionbox{CT-ICH dataset}{\includegraphics[width=0.47\linewidth]{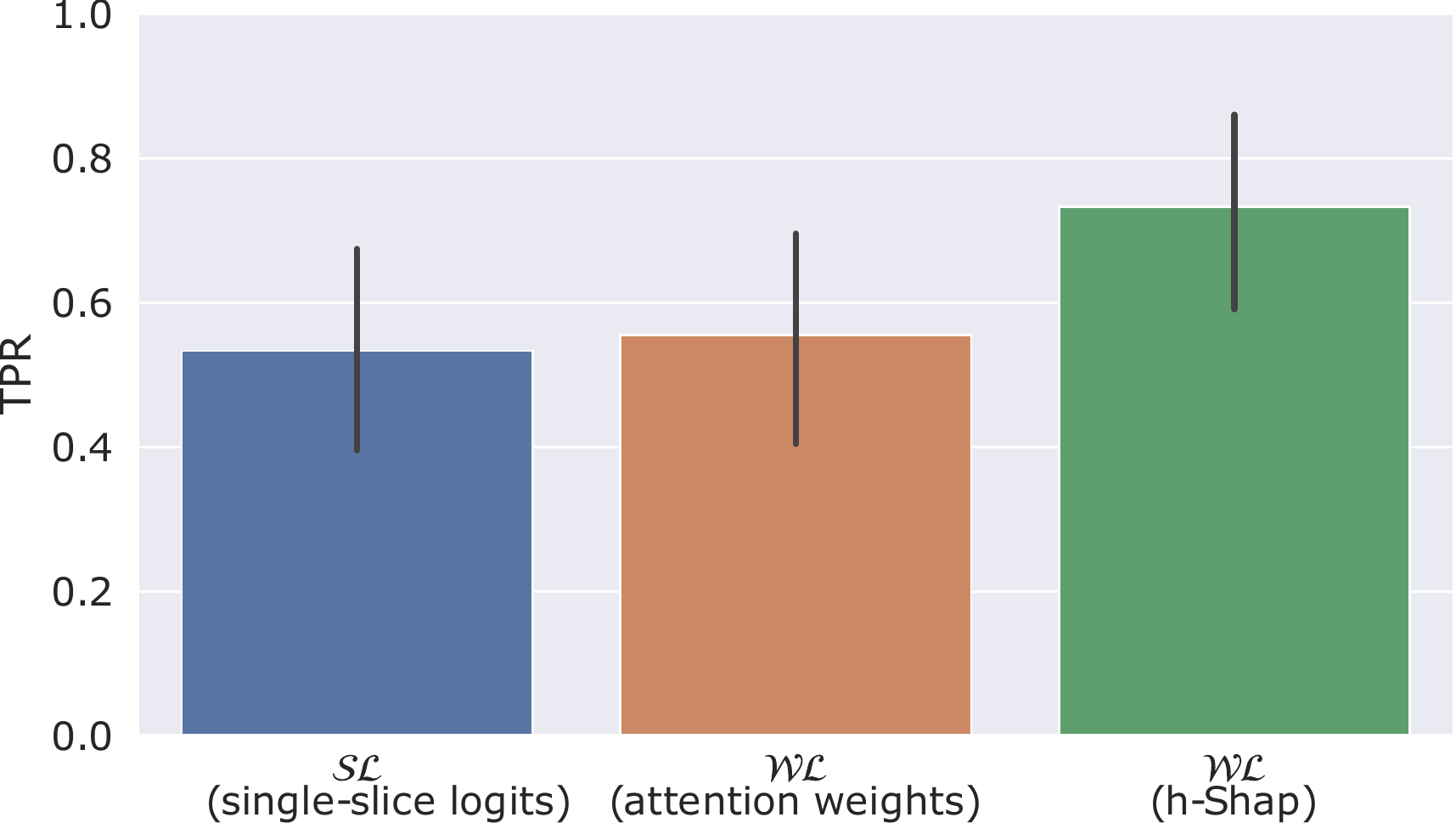}}
    \captionsetup{subrefformat=parens}
    \caption{\label{fig:exam_level_all_TPR_d}Comparison of a strong learner ($\SL$) and a weak learner ($\WL$) on examination-level hemorrhage detection. \subref{fig:exam_level_RSNA_TPR} Average recall (TPR) as a function of hemorrhage sequence length on the validation split of the RSNA dataset. \subref{fig:exam_level_CT-ICH_TPR} Average TPR on the CT-ICH dataset. These results are computed by choosing the threshold $t^*$ that minimizes the distance to the $(0,1)$ point.}
\end{figure}

\begin{figure}[h]
    \centering
    \includegraphics[width=0.75\linewidth]{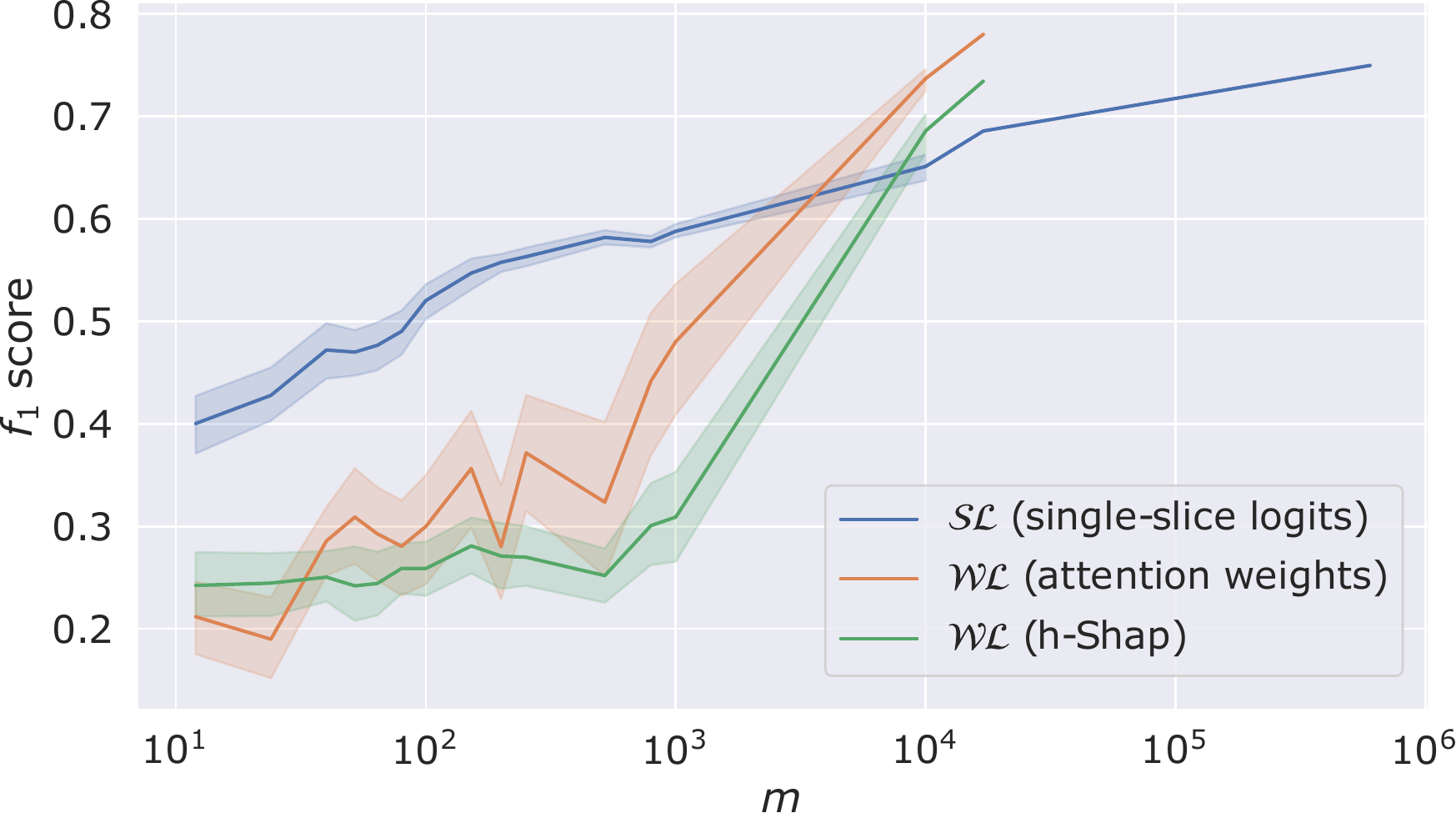}
    \caption{\label{fig:exam_level_RSNA_f1_d_label_complexity}Mean examination-level hemorrhage detection $f_1$ score as a function of number of labels $m$ on a fixed subset of 1,000 examinations from the validation split of the RSNA dataset. These results are computed by choosing the threshold $t^*$ that minimizes the distance to the $(0,1)$ point.}
\end{figure}

\newpage
\clearpage
\bibliographystyle{plainnat}
\bibliography{bibliography}

\begin{thebibliography}{95}
\providecommand{\natexlab}[1]{#1}
\providecommand{\url}[1]{\texttt{#1}}
\expandafter\ifx\csname urlstyle\endcsname\relax
  \providecommand{\doi}[1]{doi: #1}\else
  \providecommand{\doi}{doi: \begingroup \urlstyle{rm}\Url}\fi

\bibitem[Acosta et~al.(2022)Acosta, Falcone, Rajpurkar, and
  Topol]{acosta2022multimodal}
Julián~N Acosta, Guido~J Falcone, Pranav Rajpurkar, and Eric~J Topol.
\newblock Multimodal biomedical ai.
\newblock \emph{Nature Medicine}, 28\penalty0 (9):\penalty0 1773--1784,
  September 2022.

\bibitem[Ahishakiye et~al.(2021)Ahishakiye, Bastiaan Van~Gijzen, Tumwiine,
  Wario, and Obungoloch]{ahishakiye2021survey}
Emmanuel Ahishakiye, Martin Bastiaan Van~Gijzen, Julius Tumwiine, Ruth Wario,
  and Johnes Obungoloch.
\newblock A survey on deep learning in medical image reconstruction.
\newblock \emph{Intelligent Medicine}, 1\penalty0 (03):\penalty0 118--127,
  2021.

\bibitem[Alenezi and Santosh(2021)]{alenezi2021geometric}
Fayadh Alenezi and KC~Santosh.
\newblock Geometric regularized hopfield neural network for medical image
  enhancement.
\newblock \emph{International Journal of Biomedical Imaging}, 2021, 2021.

\bibitem[Amores(2013)]{amores2013multiple}
Jaume Amores.
\newblock Multiple instance classification: Review, taxonomy and comparative
  study.
\newblock \emph{Artificial intelligence}, 201:\penalty0 81--105, 2013.

\bibitem[An et~al.(2017)An, Kim, and Yoon]{an2017epidemiology}
Sang~Joon An, Tae~Jung Kim, and Byung-Woo Yoon.
\newblock Epidemiology, risk factors, and clinical features of intracerebral
  hemorrhage: an update.
\newblock \emph{Journal of stroke}, 19\penalty0 (1):\penalty0 3, 2017.

\bibitem[Andrews et~al.(2002)Andrews, Tsochantaridis, and
  Hofmann]{andrews2002support}
Stuart Andrews, Ioannis Tsochantaridis, and Thomas Hofmann.
\newblock Support vector machines for multiple-instance learning.
\newblock \emph{Advances in neural information processing systems}, 15, 2002.

\bibitem[Attia et~al.(2021)Attia, Harmon, Behr, and
  Friedman]{attia2021application}
Zachi~I Attia, David~M Harmon, Elijah~R Behr, and Paul~A Friedman.
\newblock Application of artificial intelligence to the electrocardiogram.
\newblock \emph{European heart journal}, 42\penalty0 (46):\penalty0 4717--4730,
  2021.

\bibitem[Auer et~al.(1998)Auer, Long, and Srinivasan]{auer1998approximating}
Peter Auer, Philip~M Long, and Aravind Srinivasan.
\newblock Approximating hyper-rectangles: learning and pseudorandom sets.
\newblock \emph{Journal of Computer and System Sciences}, 57\penalty0
  (3):\penalty0 376--388, 1998.

\bibitem[Bahdanau et~al.(2014)Bahdanau, Cho, and Bengio]{bahdanau2014neural}
Dzmitry Bahdanau, Kyunghyun Cho, and Yoshua Bengio.
\newblock Neural machine translation by jointly learning to align and
  translate.
\newblock \emph{arXiv preprint arXiv:1409.0473}, 2014.

\bibitem[Barnes(2015)]{barnes2015azure}
Jeff Barnes.
\newblock Azure machine learning.
\newblock In \emph{Microsoft Azure Essentials}. Microsoft, 2015.

\bibitem[Blum and Kalai(1998)]{blum1998note}
Avrim Blum and Adam Kalai.
\newblock A note on learning from multiple-instance examples.
\newblock \emph{Machine learning}, 30\penalty0 (1):\penalty0 23--29, 1998.

\bibitem[Buchlak et~al.(2022)Buchlak, Milne, Seah, Johnson, Samarasinghe,
  Hachey, Esmaili, Tran, Leveque, Farrokhi, et~al.]{buchlak2022charting}
Quinlan~D Buchlak, Michael~R Milne, Jarrel Seah, Andrew Johnson, Gihan
  Samarasinghe, Ben Hachey, Nazanin Esmaili, Aengus Tran, Jean-Christophe
  Leveque, Farrokh Farrokhi, et~al.
\newblock Charting the potential of brain computed tomography deep learning
  systems.
\newblock \emph{Journal of Clinical Neuroscience}, 99:\penalty0 217--223, 2022.

\bibitem[Burduja et~al.(2020)Burduja, Ionescu, and Verga]{burduja2020accurate}
Mihail Burduja, Radu~Tudor Ionescu, and Nicolae Verga.
\newblock Accurate and efficient intracranial hemorrhage detection and subtype
  classification in 3d ct scans with convolutional and long short-term memory
  neural networks.
\newblock \emph{Sensors}, 20\penalty0 (19):\penalty0 5611, 2020.

\bibitem[Burkart and Huber(2021)]{burkart2021survey}
Nadia Burkart and Marco~F Huber.
\newblock A survey on the explainability of supervised machine learning.
\newblock \emph{Journal of Artificial Intelligence Research}, 70:\penalty0
  245--317, 2021.

\bibitem[Buzug(2011)]{buzug2011computed}
Thorsten~M Buzug.
\newblock Computed tomography.
\newblock In \emph{Springer handbook of medical technology}, pages 311--342.
  Springer, 2011.

\bibitem[Cai et~al.(2020)Cai, Gao, and Zhao]{cai2020review}
Lei Cai, Jingyang Gao, and Di~Zhao.
\newblock A review of the application of deep learning in medical image
  classification and segmentation.
\newblock \emph{Annals of translational medicine}, 8\penalty0 (11), 2020.

\bibitem[Chen et~al.(2022)Chen, Sch{\"o}nlieb, Li{\`o}, Leiner, Dragotti, Wang,
  Rueckert, Firmin, and Yang]{chen2022ai}
Yutong Chen, Carola-Bibiane Sch{\"o}nlieb, Pietro Li{\`o}, Tim Leiner,
  Pier~Luigi Dragotti, Ge~Wang, Daniel Rueckert, David Firmin, and Guang Yang.
\newblock Ai-based reconstruction for fast mri—a systematic review and
  meta-analysis.
\newblock \emph{Proceedings of the IEEE}, 110\penalty0 (2):\penalty0 224--245,
  2022.

\bibitem[Cheplygina et~al.(2019)Cheplygina, de~Bruijne, and
  Pluim]{cheplygina2019not}
Veronika Cheplygina, Marleen de~Bruijne, and Josien~PW Pluim.
\newblock Not-so-supervised: a survey of semi-supervised, multi-instance, and
  transfer learning in medical image analysis.
\newblock \emph{Medical image analysis}, 54:\penalty0 280--296, 2019.

\bibitem[Chien et~al.(2022)Chien, Lee, Hu, and Wu]{chien2022usefulness}
Jong-Chih Chien, Jiann-Der Lee, Ching-Shu Hu, and Chieh-Tsai Wu.
\newblock The usefulness of gradient-weighted cam in assisting medical
  diagnoses.
\newblock \emph{Applied Sciences}, 12\penalty0 (15):\penalty0 7748, 2022.

\bibitem[Chilamkurthy et~al.(2018)Chilamkurthy, Ghosh, Tanamala, Biviji,
  Campeau, Venugopal, Mahajan, Rao, and Warier]{chilamkurthy2018deep}
Sasank Chilamkurthy, Rohit Ghosh, Swetha Tanamala, Mustafa Biviji, Norbert~G
  Campeau, Vasantha~Kumar Venugopal, Vidur Mahajan, Pooja Rao, and Prashant
  Warier.
\newblock Deep learning algorithms for detection of critical findings in head
  ct scans: a retrospective study.
\newblock \emph{The Lancet}, 392\penalty0 (10162):\penalty0 2388--2396, 2018.

\bibitem[Choy et~al.(2018)Choy, Khalilzadeh, Michalski, Do, Samir, Pianykh,
  Geis, Pandharipande, Brink, and Dreyer]{choy2018current}
Garry Choy, Omid Khalilzadeh, Mark Michalski, Synho Do, Anthony~E Samir, Oleg~S
  Pianykh, J~Raymond Geis, Pari~V Pandharipande, James~A Brink, and Keith~J
  Dreyer.
\newblock Current applications and future impact of machine learning in
  radiology.
\newblock \emph{Radiology}, 288\penalty0 (2):\penalty0 318, 2018.

\bibitem[Coifman and Donoho(1995)]{coifman1995translation}
Ronald~R Coifman and David~L Donoho.
\newblock Translation-invariant de-noising.
\newblock In \emph{Wavelets and statistics}, pages 125--150. Springer, 1995.

\bibitem[Correia et~al.(2019)Correia, Niculae, and
  Martins]{correia19adaptively}
Gon\c{c}alo~M Correia, Vlad Niculae, and Andr{\'e}~FT Martins.
\newblock Adaptively sparse transformers.
\newblock In \emph{Proc. EMNLP-IJCNLP (to appear)}, 2019.

\bibitem[Covert et~al.(2021)Covert, Lundberg, and Lee]{covert2021explaining}
Ian Covert, Scott Lundberg, and Su-In Lee.
\newblock Explaining by removing: A unified framework for model explanation.
\newblock \emph{Journal of Machine Learning Research}, 22\penalty0
  (209):\penalty0 1--90, 2021.

\bibitem[Covert et~al.(2022)Covert, Kim, and Lee]{covert2022learning}
Ian Covert, Chanwoo Kim, and Su-In Lee.
\newblock Learning to estimate shapley values with vision transformers.
\newblock \emph{arXiv preprint arXiv:2206.05282}, 2022.

\bibitem[Deepika et~al.(2022)Deepika, Sistla, Subramaniam, and
  Rao]{deepika2022deep}
Pon Deepika, Prasad Sistla, Ganesh Subramaniam, and Madhav Rao.
\newblock Deep learning based automated screening for intracranial hemorrhages
  and grad-cam visualizations on non-contrast head computed tomography volumes.
\newblock In \emph{2022 IEEE-EMBS International Conference on Biomedical and
  Health Informatics (BHI)}, pages 01--05. IEEE, 2022.

\bibitem[DeLong et~al.(1988)DeLong, DeLong, and
  Clarke-Pearson]{delong1988comparing}
Elizabeth~R DeLong, David~M DeLong, and Daniel~L Clarke-Pearson.
\newblock Comparing the areas under two or more correlated receiver operating
  characteristic curves: a nonparametric approach.
\newblock \emph{Biometrics}, pages 837--845, 1988.

\bibitem[Deng et~al.(2009)Deng, Dong, Socher, Li, Li, and
  Fei-Fei]{deng2009imagenet}
Jia Deng, Wei Dong, Richard Socher, Li-Jia Li, Kai Li, and Li~Fei-Fei.
\newblock Imagenet: A large-scale hierarchical image database.
\newblock In \emph{2009 IEEE conference on computer vision and pattern
  recognition}, pages 248--255. Ieee, 2009.

\bibitem[Dietterich et~al.(1997)Dietterich, Lathrop, and
  Lozano-P{\'e}rez]{dietterich1997solving}
Thomas~G Dietterich, Richard~H Lathrop, and Tom{\'a}s Lozano-P{\'e}rez.
\newblock Solving the multiple instance problem with axis-parallel rectangles.
\newblock \emph{Artificial intelligence}, 89\penalty0 (1-2):\penalty0 31--71,
  1997.

\bibitem[Dosovitskiy et~al.(2020)Dosovitskiy, Beyer, Kolesnikov, Weissenborn,
  Zhai, Unterthiner, Dehghani, Minderer, Heigold, Gelly,
  et~al.]{dosovitskiy2020image}
Alexey Dosovitskiy, Lucas Beyer, Alexander Kolesnikov, Dirk Weissenborn,
  Xiaohua Zhai, Thomas Unterthiner, Mostafa Dehghani, Matthias Minderer, Georg
  Heigold, Sylvain Gelly, et~al.
\newblock An image is worth 16x16 words: Transformers for image recognition at
  scale.
\newblock \emph{arXiv preprint arXiv:2010.11929}, 2020.

\bibitem[Ethayarajh and Jurafsky(2021)]{ethayarajh2021attention}
Kawin Ethayarajh and Dan Jurafsky.
\newblock Attention flows are shapley value explanations.
\newblock \emph{arXiv preprint arXiv:2105.14652}, 2021.

\bibitem[Eyuboglu et~al.(2021)Eyuboglu, Angus, Patel, Pareek, Davidzon, Long,
  Dunnmon, and Lungren]{eyuboglu2021multi}
Sabri Eyuboglu, Geoffrey Angus, Bhavik~N Patel, Anuj Pareek, Guido Davidzon,
  Jin Long, Jared Dunnmon, and Matthew~P Lungren.
\newblock Multi-task weak supervision enables anatomically-resolved abnormality
  detection in whole-body fdg-pet/ct.
\newblock \emph{Nature communications}, 12\penalty0 (1):\penalty0 1--15, 2021.

\bibitem[Flanders et~al.(2020)Flanders, Prevedello, Shih, Halabi,
  Kalpathy-Cramer, Ball, Mongan, Stein, Kitamura, Lungren,
  et~al.]{flanders2020construction}
Adam~E Flanders, Luciano~M Prevedello, George Shih, Safwan~S Halabi, Jayashree
  Kalpathy-Cramer, Robyn Ball, John~T Mongan, Anouk Stein, Felipe~C Kitamura,
  Matthew~P Lungren, et~al.
\newblock Construction of a machine learning dataset through collaboration: the
  rsna 2019 brain ct hemorrhage challenge.
\newblock \emph{Radiology: Artificial Intelligence}, 2\penalty0 (3):\penalty0
  e190211, 2020.

\bibitem[Gaube et~al.(2021)Gaube, Suresh, Raue, Merritt, Berkowitz, Lermer,
  Coughlin, Guttag, Colak, and Ghassemi]{Gaube2021DoAsAiSay}
Susanne Gaube, Harini Suresh, Martina Raue, Alexander Merritt, Seth~J
  Berkowitz, Eva Lermer, Joseph~F Coughlin, John~V Guttag, Errol Colak, and
  Marzyeh Ghassemi.
\newblock {Do as AI say: susceptibility in deployment of clinical
  decision-aids}.
\newblock \emph{npj Digital Medicine}, 4\penalty0 (1):\penalty0 31, 2021.
\newblock ISSN 2398-6352.
\newblock \doi{10.1038/s41746-021-00385-9}.
\newblock URL \url{https://doi.org/10.1038/s41746-021-00385-9}.

\bibitem[Giger(2018)]{giger2018machine}
Maryellen~L Giger.
\newblock Machine learning in medical imaging.
\newblock \emph{Journal of the American College of Radiology}, 15\penalty0
  (3):\penalty0 512--520, 2018.

\bibitem[Goldberger et~al.(2000)Goldberger, Amaral, Glass, Hausdorff, Ivanov,
  Mark, Mietus, Moody, Peng, and Stanley]{goldberger2000physiobank}
Ary~L Goldberger, Luis~AN Amaral, Leon Glass, Jeffrey~M Hausdorff, Plamen~Ch
  Ivanov, Roger~G Mark, Joseph~E Mietus, George~B Moody, Chung-Kang Peng, and
  H~Eugene Stanley.
\newblock Physiobank, physiotoolkit, and physionet: components of a new
  research resource for complex physiologic signals.
\newblock \emph{circulation}, 101\penalty0 (23):\penalty0 e215--e220, 2000.

\bibitem[Goodman and Flaxman(2017)]{goodman2017european}
Bryce Goodman and Seth Flaxman.
\newblock European union regulations on algorithmic decision-making and a
  “right to explanation”.
\newblock \emph{AI magazine}, 38\penalty0 (3):\penalty0 50--57, 2017.

\bibitem[Gudigar et~al.(2021)Gudigar, Raghavendra, Hegde, Menon, Molinari,
  Ciaccio, and Acharya]{gudigar2021automated}
Anjan Gudigar, U~Raghavendra, Ajay Hegde, Girish~R Menon, Filippo Molinari,
  Edward~J Ciaccio, and U~Rajendra Acharya.
\newblock Automated detection and screening of traumatic brain injury (tbi)
  using computed tomography images: a comprehensive review and future
  perspectives.
\newblock \emph{International journal of environmental research and public
  health}, 18\penalty0 (12):\penalty0 6499, 2021.

\bibitem[Habibzadeh et~al.(2016)Habibzadeh, Habibzadeh, and
  Yadollahie]{habibzadeh2016determining}
Farrokh Habibzadeh, Parham Habibzadeh, and Mahboobeh Yadollahie.
\newblock On determining the most appropriate test cut-off value: the case of
  tests with continuous results.
\newblock \emph{Biochemia medica}, 26\penalty0 (3):\penalty0 297--307, 2016.

\bibitem[He et~al.(2016)He, Zhang, Ren, and Sun]{he2016deep}
Kaiming He, Xiangyu Zhang, Shaoqing Ren, and Jian Sun.
\newblock Deep residual learning for image recognition.
\newblock In \emph{Proceedings of the IEEE conference on computer vision and
  pattern recognition}, pages 770--778, 2016.

\bibitem[Hssayeni(2020)]{hssayeni2020computed}
Murtadha Hssayeni.
\newblock Computed tomography images for intracranial hemorrhage detection and
  segmentation.
\newblock \emph{Intracranial Hemorrhage Segmentation Using A Deep Convolutional
  Model. Data}, 5\penalty0 (1), 2020.

\bibitem[Hssayeni et~al.(2020)Hssayeni, Croock, Salman, Al-khafaji, Yahya, and
  Ghoraani]{hssayeni2020intracranial}
Murtadha~D Hssayeni, Muayad~S Croock, Aymen~D Salman, Hassan~Falah Al-khafaji,
  Zakaria~A Yahya, and Behnaz Ghoraani.
\newblock Intracranial hemorrhage segmentation using a deep convolutional
  model.
\newblock \emph{Data}, 5\penalty0 (1):\penalty0 14, 2020.

\bibitem[Ilse et~al.(2018)Ilse, Tomczak, and Welling]{ilse2018attention}
Maximilian Ilse, Jakub Tomczak, and Max Welling.
\newblock Attention-based deep multiple instance learning.
\newblock In \emph{International conference on machine learning}, pages
  2127--2136. PMLR, 2018.

\bibitem[Jain et~al.(2022)Jain, Salman, Wong, Zhang, Vineet, Vemprala, and
  Madry]{jain2022missingness}
Saachi Jain, Hadi Salman, Eric Wong, Pengchuan Zhang, Vibhav Vineet, Sai
  Vemprala, and Aleksander Madry.
\newblock Missingness bias in model debugging.
\newblock \emph{arXiv preprint arXiv:2204.08945}, 2022.

\bibitem[Jain and Wallace(2019)]{jain2019attention}
Sarthak Jain and Byron~C Wallace.
\newblock Attention is not explanation.
\newblock \emph{arXiv preprint arXiv:1902.10186}, 2019.

\bibitem[Kaka et~al.(2021)Kaka, Zhang, and Khan]{kaka2021artificial}
Hussam Kaka, Euan Zhang, and Nazir Khan.
\newblock Artificial intelligence and deep learning in neuroradiology:
  exploring the new frontier.
\newblock \emph{Canadian Association of Radiologists Journal}, 72\penalty0
  (1):\penalty0 35--44, 2021.

\bibitem[Kang et~al.(2017)Kang, Min, and Ye]{kang2017deep}
Eunhee Kang, Junhong Min, and Jong~Chul Ye.
\newblock A deep convolutional neural network using directional wavelets for
  low-dose x-ray ct reconstruction.
\newblock \emph{Medical physics}, 44\penalty0 (10):\penalty0 e360--e375, 2017.

\bibitem[Kawooya(2012)]{Kawooya2012-il}
Michael~G Kawooya.
\newblock Training for rural radiology and imaging in {Sub-Saharan} africa:
  Addressing the mismatch between services and population.
\newblock \emph{J. Clin. Imaging Sci.}, 2, 2012.

\bibitem[Khan et~al.(2022)Khan, Naseer, Hayat, Zamir, Khan, and
  Shah]{khan2022transformers}
Salman Khan, Muzammal Naseer, Munawar Hayat, Syed~Waqas Zamir, Fahad~Shahbaz
  Khan, and Mubarak Shah.
\newblock Transformers in vision: A survey.
\newblock \emph{ACM computing surveys (CSUR)}, 54\penalty0 (10s):\penalty0
  1--41, 2022.

\bibitem[Kingma and Ba(2014)]{kingma2014adam}
Diederik~P Kingma and Jimmy Ba.
\newblock Adam: A method for stochastic optimization.
\newblock \emph{arXiv preprint arXiv:1412.6980}, 2014.

\bibitem[Langlotz(2019)]{langlotz2019will}
Curtis~P Langlotz.
\newblock Will artificial intelligence replace radiologists?
\newblock \emph{Radiology. Artificial intelligence}, 1\penalty0 (3), 2019.

\bibitem[Latif et~al.(2019)Latif, Xiao, Imran, and Tu]{latif2019medical}
Jahanzaib Latif, Chuangbai Xiao, Azhar Imran, and Shanshan Tu.
\newblock Medical imaging using machine learning and deep learning algorithms:
  a review.
\newblock In \emph{2019 2nd International conference on computing, mathematics
  and engineering technologies (iCoMET)}, pages 1--5. IEEE, 2019.

\bibitem[Lee et~al.(2019)Lee, Yune, Mansouri, Kim, Tajmir, Guerrier, Ebert,
  Pomerantz, Romero, Kamalian, et~al.]{lee2019explainable}
Hyunkwang Lee, Sehyo Yune, Mohammad Mansouri, Myeongchan Kim, Shahein~H Tajmir,
  Claude~E Guerrier, Sarah~A Ebert, Stuart~R Pomerantz, Javier~M Romero,
  Shahmir Kamalian, et~al.
\newblock An explainable deep-learning algorithm for the detection of acute
  intracranial haemorrhage from small datasets.
\newblock \emph{Nature biomedical engineering}, 3\penalty0 (3):\penalty0
  173--182, 2019.

\bibitem[Lee et~al.(2020)Lee, Kim, Kim, and Kim]{lee2020detection}
Ji~Young Lee, Jong~Soo Kim, Tae~Yoon Kim, and Young~Soo Kim.
\newblock Detection and classification of intracranial haemorrhage on ct images
  using a novel deep-learning algorithm.
\newblock \emph{Scientific Reports}, 10\penalty0 (1):\penalty0 1--7, 2020.

\bibitem[Lin et~al.(2017)Lin, Goyal, Girshick, He, and
  Doll{\'a}r]{lin2017focal}
Tsung-Yi Lin, Priya Goyal, Ross Girshick, Kaiming He, and Piotr Doll{\'a}r.
\newblock Focal loss for dense object detection.
\newblock In \emph{Proceedings of the IEEE international conference on computer
  vision}, pages 2980--2988, 2017.

\bibitem[Linardatos et~al.(2020)Linardatos, Papastefanopoulos, and
  Kotsiantis]{linardatos2020explainable}
Pantelis Linardatos, Vasilis Papastefanopoulos, and Sotiris Kotsiantis.
\newblock Explainable ai: A review of machine learning interpretability
  methods.
\newblock \emph{Entropy}, 23\penalty0 (1):\penalty0 18, 2020.

\bibitem[L{\'o}pez-P{\'e}rez et~al.(2022)L{\'o}pez-P{\'e}rez, Schmidt, Wu,
  Molina, and Katsaggelos]{lopez2022deep}
Miguel L{\'o}pez-P{\'e}rez, Arne Schmidt, Yunan Wu, Rafael Molina, and
  Aggelos~K Katsaggelos.
\newblock Deep gaussian processes for multiple instance learning: Application
  to ct intracranial hemorrhage detection.
\newblock \emph{Computer Methods and Programs in Biomedicine}, 219:\penalty0
  106783, 2022.

\bibitem[Lundberg and Lee(2017)]{lundberg2017unified}
Scott~M Lundberg and Su-In Lee.
\newblock A unified approach to interpreting model predictions.
\newblock \emph{Advances in neural information processing systems}, 30, 2017.

\bibitem[Maron and Lozano-P{\'e}rez(1997)]{maron1997framework}
Oded Maron and Tom{\'a}s Lozano-P{\'e}rez.
\newblock A framework for multiple-instance learning.
\newblock \emph{Advances in neural information processing systems}, 10, 1997.

\bibitem[Martins and Astudillo(2016)]{martins2016softmax}
Andre Martins and Ramon Astudillo.
\newblock From softmax to sparsemax: A sparse model of attention and
  multi-label classification.
\newblock In \emph{International conference on machine learning}, pages
  1614--1623. PMLR, 2016.

\bibitem[Montagnon et~al.(2020)Montagnon, Cerny, Cadrin-Ch{\^e}nevert,
  Hamilton, Derennes, Ilinca, Vandenbroucke-Menu, Turcotte, Kadoury, and
  Tang]{montagnon2020deep}
Emmanuel Montagnon, Milena Cerny, Alexandre Cadrin-Ch{\^e}nevert, Vincent
  Hamilton, Thomas Derennes, Andr{\'e} Ilinca, Franck Vandenbroucke-Menu, Simon
  Turcotte, Samuel Kadoury, and An~Tang.
\newblock Deep learning workflow in radiology: a primer.
\newblock \emph{Insights into imaging}, 11\penalty0 (1):\penalty0 1--15, 2020.

\bibitem[Panesar and Panesar(2020)]{panesar2020artificial}
Arjun Panesar and Harkrishan Panesar.
\newblock Artificial intelligence and machine learning in global healthcare.
\newblock \emph{Handbook of Global Health}, pages 1--39, 2020.

\bibitem[Panwar et~al.(2020)Panwar, Gupta, Siddiqui, Morales-Menendez,
  Bhardwaj, and Singh]{panwar2020deep}
Harsh Panwar, PK~Gupta, Mohammad~Khubeb Siddiqui, Ruben Morales-Menendez,
  Prakhar Bhardwaj, and Vaishnavi Singh.
\newblock A deep learning and grad-cam based color visualization approach for
  fast detection of covid-19 cases using chest x-ray and ct-scan images.
\newblock \emph{Chaos, Solitons \& Fractals}, 140:\penalty0 110190, 2020.

\bibitem[Patel et~al.(2019)Patel, Rosenberg, Willcox, Baltaxe, Lyons, Irvin,
  Rajpurkar, Amrhein, Gupta, Halabi, Langlotz, Lo, Mammarappallil, Mariano,
  Riley, Seekins, Shen, Zucker, and Lungren]{Patel2019-vk}
Bhavik~N Patel, Louis Rosenberg, Gregg Willcox, David Baltaxe, Mimi Lyons,
  Jeremy Irvin, Pranav Rajpurkar, Timothy Amrhein, Rajan Gupta, Safwan Halabi,
  Curtis Langlotz, Edward Lo, Joseph Mammarappallil, A~J Mariano, Geoffrey
  Riley, Jayne Seekins, Luyao Shen, Evan Zucker, and Matthew Lungren.
\newblock Human-machine partnership with artificial intelligence for chest
  radiograph diagnosis.
\newblock \emph{NPJ Digit Med}, 2:\penalty0 111, November 2019.

\bibitem[P{\'e}rez-Garc{\'\i}a et~al.(2021)P{\'e}rez-Garc{\'\i}a, Sparks, and
  Ourselin]{perez2021torchio}
Fernando P{\'e}rez-Garc{\'\i}a, Rachel Sparks, and Sebastien Ourselin.
\newblock Torchio: a python library for efficient loading, preprocessing,
  augmentation and patch-based sampling of medical images in deep learning.
\newblock \emph{Computer Methods and Programs in Biomedicine}, 208:\penalty0
  106236, 2021.

\bibitem[Perkins and Schisterman(2006)]{perkins2006inconsistency}
Neil~J Perkins and Enrique~F Schisterman.
\newblock The inconsistency of “optimal” cutpoints obtained using two
  criteria based on the receiver operating characteristic curve.
\newblock \emph{American journal of epidemiology}, 163\penalty0 (7):\penalty0
  670--675, 2006.

\bibitem[Peters et~al.(2019)Peters, Niculae, and Martins]{peters2019sparse}
Ben Peters, Vlad Niculae, and Andr{\'e} F.~T. Martins.
\newblock Sparse sequence-to-sequence models.
\newblock In \emph{Proceedings of the 57th Annual Meeting of the Association
  for Computational Linguistics}, pages 1504--1519, Florence, Italy, July 2019.
  Association for Computational Linguistics.
\newblock \doi{10.18653/v1/P19-1146}.
\newblock URL \url{https://aclanthology.org/P19-1146}.

\bibitem[Quellec et~al.(2017)Quellec, Cazuguel, Cochener, and
  Lamard]{quellec2017multiple}
Gwenol{\'e} Quellec, Guy Cazuguel, B{\'e}atrice Cochener, and Mathieu Lamard.
\newblock Multiple-instance learning for medical image and video analysis.
\newblock \emph{IEEE reviews in biomedical engineering}, 10:\penalty0 213--234,
  2017.

\bibitem[Rajpurkar et~al.(2018)Rajpurkar, Irvin, Ball, Zhu, Yang, Mehta, Duan,
  Ding, Bagul, Langlotz, Patel, Yeom, Shpanskaya, Blankenberg, Seekins,
  Amrhein, Mong, Halabi, Zucker, Ng, and Lungren]{Rajpurkar2018-pj}
Pranav Rajpurkar, Jeremy Irvin, Robyn~L Ball, Kaylie Zhu, Brandon Yang, Hershel
  Mehta, Tony Duan, Daisy Ding, Aarti Bagul, Curtis~P Langlotz, Bhavik~N Patel,
  Kristen~W Yeom, Katie Shpanskaya, Francis~G Blankenberg, Jayne Seekins,
  Timothy~J Amrhein, David~A Mong, Safwan~S Halabi, Evan~J Zucker, Andrew~Y Ng,
  and Matthew~P Lungren.
\newblock Deep learning for chest radiograph diagnosis: A retrospective
  comparison of the {CheXNeXt} algorithm to practicing radiologists.
\newblock \emph{PLoS Med.}, 15\penalty0 (11):\penalty0 e1002686, November 2018.

\bibitem[Reis et~al.(2020)Reis, Nascimento, Aranha, Secol, Machado, Felix,
  Stein, and Amaro]{reis2020brain}
Eduardo~Pontes Reis, Felipe Nascimento, Mateus Aranha, F~Mainetti Secol,
  Birajara Machado, Marcelo Felix, Anouk Stein, and Edson Amaro.
\newblock Brain hemorrhage extended (bhx): Bounding box extrapolation from
  thick to thin slice ct images, 2020.

\bibitem[Roscher et~al.(2020)Roscher, Bohn, Duarte, and
  Garcke]{roscher2020explainable}
Ribana Roscher, Bastian Bohn, Marco~F Duarte, and Jochen Garcke.
\newblock Explainable machine learning for scientific insights and discoveries.
\newblock \emph{Ieee Access}, 8:\penalty0 42200--42216, 2020.

\bibitem[Saab et~al.(2019)Saab, Dunnmon, Goldman, Ratner, Sagreiya, R{\'e}, and
  Rubin]{saab2019doubly}
Khaled Saab, Jared Dunnmon, Roger Goldman, Alex Ratner, Hersh Sagreiya,
  Christopher R{\'e}, and Daniel Rubin.
\newblock Doubly weak supervision of deep learning models for head ct.
\newblock In \emph{International Conference on Medical Image Computing and
  Computer-Assisted Intervention}, pages 811--819. Springer, 2019.

\bibitem[Saba et~al.(2019)Saba, Biswas, Kuppili, Godia, Suri, Edla, Omerzu,
  Laird, Khanna, Mavrogeni, et~al.]{saba2019present}
Luca Saba, Mainak Biswas, Venkatanareshbabu Kuppili, Elisa~Cuadrado Godia,
  Harman~S Suri, Damodar~Reddy Edla, Toma{\v{z}} Omerzu, John~R Laird,
  Narendra~N Khanna, Sophie Mavrogeni, et~al.
\newblock The present and future of deep learning in radiology.
\newblock \emph{European journal of radiology}, 114:\penalty0 14--24, 2019.

\bibitem[Sabato and Tishby(2009)]{sabato2009homogeneous}
Sivan Sabato and Naftali Tishby.
\newblock Homogeneous multi-instance learning with arbitrary dependence.
\newblock In \emph{COLT}. Citeseer, 2009.

\bibitem[Sabato and Tishby(2012)]{sabato2012multi}
Sivan Sabato and Naftali Tishby.
\newblock Multi-instance learning with any hypothesis class.
\newblock \emph{The Journal of Machine Learning Research}, 13\penalty0
  (1):\penalty0 2999--3039, 2012.

\bibitem[Sabato et~al.(2010)Sabato, Srebro, and Tishby]{sabato2010reducing}
Sivan Sabato, Nathan Srebro, and Naftali Tishby.
\newblock Reducing label complexity by learning from bags.
\newblock In \emph{Proceedings of the Thirteenth International Conference on
  Artificial Intelligence and Statistics}, pages 685--692. JMLR Workshop and
  Conference Proceedings, 2010.

\bibitem[Schlemper et~al.(2019)Schlemper, Oktay, Schaap, Heinrich, Kainz,
  Glocker, and Rueckert]{schlemper2019attention}
Jo~Schlemper, Ozan Oktay, Michiel Schaap, Mattias Heinrich, Bernhard Kainz, Ben
  Glocker, and Daniel Rueckert.
\newblock Attention gated networks: Learning to leverage salient regions in
  medical images.
\newblock \emph{Medical image analysis}, 53:\penalty0 197--207, 2019.

\bibitem[Selvaraju et~al.(2017)Selvaraju, Cogswell, Das, Vedantam, Parikh, and
  Batra]{selvaraju2017grad}
Ramprasaath~R Selvaraju, Michael Cogswell, Abhishek Das, Ramakrishna Vedantam,
  Devi Parikh, and Dhruv Batra.
\newblock Grad-cam: Visual explanations from deep networks via gradient-based
  localization.
\newblock In \emph{Proceedings of the IEEE international conference on computer
  vision}, pages 618--626, 2017.

\bibitem[Shapley(1953)]{Shapley1953}
Lloyd~S Shapley.
\newblock {A value for n-person games}.
\newblock \emph{Contributions to the Theory of Games}, 2\penalty0
  (28):\penalty0 307--317, 1953.

\bibitem[Shehab et~al.(2022)Shehab, Abualigah, Shambour, Abu-Hashem, Shambour,
  Alsalibi, and Gandomi]{shehab2022machine}
Mohammad Shehab, Laith Abualigah, Qusai Shambour, Muhannad~A Abu-Hashem, Mohd
  Khaled~Yousef Shambour, Ahmed~Izzat Alsalibi, and Amir~H Gandomi.
\newblock Machine learning in medical applications: A review of
  state-of-the-art methods.
\newblock \emph{Computers in Biology and Medicine}, 145:\penalty0 105458, 2022.

\bibitem[Sutskever et~al.(2013)Sutskever, Martens, Dahl, and
  Hinton]{sutskever2013importance}
Ilya Sutskever, James Martens, George Dahl, and Geoffrey Hinton.
\newblock On the importance of initialization and momentum in deep learning.
\newblock In \emph{International conference on machine learning}, pages
  1139--1147. PMLR, 2013.

\bibitem[Teneggi et~al.(2022)Teneggi, Luster, and Sulam]{teneggi2022fast}
Jacopo Teneggi, Alexandre Luster, and Jeremias Sulam.
\newblock Fast hierarchical games for image explanations.
\newblock \emph{IEEE Transactions on Pattern Analysis and Machine
  Intelligence}, 2022.

\bibitem[Turner and Holdsworth(2011)]{turner2011ct}
PJ~Turner and G~Holdsworth.
\newblock Ct stroke window settings: an unfortunate misleading misnomer?
\newblock \emph{The British journal of radiology}, 84\penalty0 (1008):\penalty0
  1061--1066, 2011.

\bibitem[Tushar et~al.(2021)Tushar, D’Anniballe, Hou, Mazurowski, Fu, Samei,
  Rubin, and Lo]{tushar2021classification}
Fakrul~Islam Tushar, Vincent~M D’Anniballe, Rui Hou, Maciej~A Mazurowski,
  Wanyi Fu, Ehsan Samei, Geoffrey~D Rubin, and Joseph~Y Lo.
\newblock Classification of multiple diseases on body ct scans using weakly
  supervised deep learning.
\newblock \emph{Radiology: Artificial Intelligence}, 4\penalty0 (1):\penalty0
  e210026, 2021.

\bibitem[Ueda et~al.(2019)Ueda, Shimazaki, and Miki]{ueda2019technical}
Daiju Ueda, Akitoshi Shimazaki, and Yukio Miki.
\newblock Technical and clinical overview of deep learning in radiology.
\newblock \emph{Japanese journal of radiology}, 37\penalty0 (1):\penalty0
  15--33, 2019.

\bibitem[Vaswani et~al.(2017)Vaswani, Shazeer, Parmar, Uszkoreit, Jones, Gomez,
  Kaiser, and Polosukhin]{vaswani2017attention}
Ashish Vaswani, Noam Shazeer, Niki Parmar, Jakob Uszkoreit, Llion Jones,
  Aidan~N Gomez, {\L}ukasz Kaiser, and Illia Polosukhin.
\newblock Attention is all you need.
\newblock \emph{Advances in neural information processing systems}, 30, 2017.

\bibitem[Wang et~al.(2021)Wang, Lei, Fu, Wynne, Curran, Liu, and
  Yang]{wang2021review}
Tonghe Wang, Yang Lei, Yabo Fu, Jacob~F Wynne, Walter~J Curran, Tian Liu, and
  Xiaofeng Yang.
\newblock A review on medical imaging synthesis using deep learning and its
  clinical applications.
\newblock \emph{Journal of applied clinical medical physics}, 22\penalty0
  (1):\penalty0 11--36, 2021.

\bibitem[Wang et~al.(2022)Wang, Saoud, Wangsiricharoen, James, Popel, and
  Sulam]{Wang_MIL_2022}
Zhenzhen Wang, Carla Saoud, Sintawat Wangsiricharoen, Aaron~W. James,
  Aleksander~S. Popel, and Jeremias Sulam.
\newblock Label cleaning multiple instance learning: Refining coarse
  annotations on single whole-slide images.
\newblock \emph{IEEE Transactions on Medical Imaging}, pages 1--1, 2022.
\newblock \doi{10.1109/TMI.2022.3202759}.

\bibitem[Weidmann et~al.(2003)Weidmann, Frank, and Pfahringer]{weidmann2003two}
Nils Weidmann, Eibe Frank, and Bernhard Pfahringer.
\newblock A two-level learning method for generalized multi-instance problems.
\newblock In \emph{European Conference on Machine Learning}, pages 468--479.
  Springer, 2003.

\bibitem[Weissglass(2021)]{weissglass2021contextual}
Daniel~E Weissglass.
\newblock Contextual bias, the democratization of healthcare, and medical
  artificial intelligence in low-and middle-income countries.
\newblock \emph{Bioethics}, 2021.

\bibitem[Wu et~al.(2021)Wu, Schmidt, Hern{\'a}ndez-S{\'a}nchez, Molina, and
  Katsaggelos]{wu2021combining}
Yunan Wu, Arne Schmidt, Enrique Hern{\'a}ndez-S{\'a}nchez, Rafael Molina, and
  Aggelos~K Katsaggelos.
\newblock Combining attention-based multiple instance learning and gaussian
  processes for ct hemorrhage detection.
\newblock In \emph{International Conference on Medical Image Computing and
  Computer-Assisted Intervention}, pages 582--591. Springer, 2021.

\bibitem[Yeo et~al.(2021)Yeo, Tahayori, Kok, Maingard, Kutaiba, Russell, Thijs,
  Jhamb, Chandra, Brooks, et~al.]{yeo2021review}
Melissa Yeo, Bahman Tahayori, Hong~Kuan Kok, Julian Maingard, Numan Kutaiba,
  Jeremy Russell, Vincent Thijs, Ashu Jhamb, Ronil~V Chandra, Mark Brooks,
  et~al.
\newblock Review of deep learning algorithms for the automatic detection of
  intracranial hemorrhages on computed tomography head imaging.
\newblock \emph{Journal of neurointerventional surgery}, 13\penalty0
  (4):\penalty0 369--378, 2021.

\bibitem[Youden(1950)]{youden1950index}
William~J Youden.
\newblock Index for rating diagnostic tests.
\newblock \emph{Cancer}, 3\penalty0 (1):\penalty0 32--35, 1950.

\bibitem[Zaheer et~al.(2017)Zaheer, Kottur, Ravanbakhsh, Poczos, Salakhutdinov,
  and Smola]{zaheer2017deep}
Manzil Zaheer, Satwik Kottur, Siamak Ravanbakhsh, Barnabas Poczos, Russ~R
  Salakhutdinov, and Alexander~J Smola.
\newblock Deep sets.
\newblock \emph{Advances in neural information processing systems}, 30, 2017.

\bibitem[Zech et~al.(2018)Zech, Badgeley, Liu, Costa, Titano, and
  Oermann]{zech2018generalization}
John~R. Zech, Marcus~A. Badgeley, Manway Liu, Anthony~B. Costa, Joseph~J.
  Titano, and Eric~Karl Oermann.
\newblock Variable generalization performance of a deep learning model to
  detect pneumonia in chest radiographs: A cross-sectional study.
\newblock \emph{PLOS Medicine}, 15\penalty0 (11):\penalty0 1--17, 11 2018.
\newblock \doi{10.1371/journal.pmed.1002683}.
\newblock URL \url{https://doi.org/10.1371/journal.pmed.1002683}.

\end{thebibliography}

\end{document}